\documentclass{article}

\usepackage{arxiv}

\usepackage[utf8]{inputenc} 
\usepackage[T1]{fontenc}    
\usepackage{hyperref}       
\usepackage{url}            
\usepackage{booktabs}       
\usepackage{amsfonts}       
\usepackage{nicefrac}       
\usepackage{microtype}      
\usepackage{lipsum}
\usepackage{multirow} 
\newcommand{\isucaption}{\caption}
\usepackage[ruled,vlined]{algorithm2e}
\SetKw{Return}{Return}
\usepackage{float} 
\usepackage{graphicx}
\usepackage{amsmath} 
\usepackage{float}
\usepackage{xcolor}
\usepackage{amsmath}

\graphicspath{ {./images/} }

\title{APQF: Agentic Profiling-Guided Structured Pruning and Mixed-Precision Quantization with Adaptive Fine-Tuning}

\author{
Sadegh Jafari \\
Department of Computer Science \\
Iowa State University \\
Ames, IA, USA \\
\texttt{sadegh@iastate.edu}
\And
Mohiuddin Bilwal \\
Department of Computer Science \\
Iowa State University \\
Ames, IA, USA \\
\texttt{moh596@iastate.edu}
\And
Fan Zhou \\
Department of Computer Science \\
Iowa State University \\
Ames, IA, USA \\
\texttt{fanzhou@iastate.edu}
\And
Brian Gelder \\
Department of Agricultural and Biosystems Engineering \\
Iowa State University \\
Ames, IA, USA \\
\texttt{bkgelder@iastate.edu}
\And
Ali Jannesari \\
Department of Computer Science \\
Iowa State University \\
Ames, IA, USA \\
\texttt{jannesar@iastate.edu}
}

\begin{document}
\maketitle
\begin{abstract}

Modern deep neural networks achieve strong performance across many domains, but their scale makes them costly and slow, especially on resource-constrained edge devices. Pruning and quantization address this, but they rely on manual, expertise-driven choices, and their algorithms are often complex and hard to apply across different architectures. Uniform settings also ignore how differently individual modules respond to compression, which causes large accuracy loss. 

We introduce APQF, an agentic profiling-guided framework that combines structured pruning, mixed-precision quantization-aware training, and accuracy recovery in one automated pipeline. A profiling agent measures how cost is distributed across the model and how sensitive each part is to pruning. This evidence then drives a pruning agent that sets per-layer pruning ratios and a quantization agent that assigns per-layer bit-widths, while a fine-tuning agent recovers accuracy according to how much was lost and a separate evaluation agent measures every model under the same protocol. APQF is the first framework to combine LLM-guided, profiling-grounded decisions with a fully training-aware pipeline of structured pruning and mixed-precision quantization-aware training with recovery, across both convolutional networks and vision transformers. 

We evaluate APQF on ResNet, VGG7, ViT, DeiT, and Swin using ImageNet-1k and CIFAR-10. On ImageNet-1K it reduces compute to between 5.6 and 7.7 percent of the original bit-operations, a 13 to 18 times reduction, while keeping accuracy close to the baseline, and under a 200K-image training budget it stays roughly 17 points
higher in Top-1 accuracy than existing joint pruning and quantization methods. On CIFAR-10 it compresses further than the competing joint method on four of five architectures, and DeiT-Tiny, ResNet-50, and Swin-Tiny end up more accurate than their uncompressed originals. On VGG7, it reaches 93.15 percent accuracy using only 0.41 percent of the baseline bit-operations, the only method at that level of compression to improve on its full-precision baseline. An ablation shows that uniform compression loses the most accuracy at matched compute, while withholding the measured profiling data from the LLM planner reduces accuracy on every model tested. On Swin-Tiny, six LLM planners, including free open-weight ones, all reach 97.4 to 97.9 percent accuracy, so the framework does not depend on a premium model.
\end{abstract}


\section{Introduction}
Modern deep neural networks (DNNs) are used in many applications across different
domains \cite{he2016deep,devlin2019bert}. Their strong performance has made it possible to speed up tasks that previously required manual work and specialized
expertise \cite{rangarajan2025identification}. Computer vision architectures, including both CNNs and ViTs, have also been widely used in recent years, in areas ranging from agriculture \cite{kamilaris2018deep,rangarajan2025identification} to medicine \cite{litjens2017survey,abou2023white}. One of the main challenges with these models is that, despite their strong performance, their growing scale imposes substantial computational and memory costs \cite{li2023model}. This makes it difficult to run them on everyday edge devices, or even on ordinary computers. In addition, larger models take longer to produce a prediction, and this added
inference time causes delays in applications where users expect a fast
response~\cite{shuvo2022efficient}. It is therefore important to make these models
smaller and faster while keeping their performance close to that of the original
model. This is the goal of model compression. The most widely used model compression techniques are pruning, which removes redundant weights or structures from the network \cite{li2023model}; quantization, which represents weights and activations at lower numerical precision~\cite{zhou2017incremental}; and knowledge
distillation, which transfers the behavior of a large teacher model into a smaller
student \cite{hinton2015distilling}. Applying these techniques well is not
straightforward, because each architecture distributes its computational cost and
its redundancy differently. A good compression strategy therefore has to identify
which parts of a given model are redundant and which are sensitive to the task,
compress the redundant parts, and then recover the accuracy lost in the process,
so that the compressed model stays as close as possible to the original.

\subsection{Problem Statement}
\subsubsection{Manual and Expertise-Driven Compression}
A first difficulty lies in deciding \emph{how} to compress a given network. Many
pruning and quantization methods rely on hand-designed heuristics to decide what to remove or how aggressively to quantize, such as ranking weights or filters by their magnitude or norm~\cite{han2015learning,filters2016pruning}. Choosing and tuning these criteria, along with the per-layer pruning ratios and bit-widths, requires expert knowledge and considerable manual effort, and the choices that work for one model rarely transfer to another~\cite{cheng2024survey}. This difficulty is compounded by the fact that different architectures organize their computation and their internal dependencies in very different ways, so building a method that is model-agnostic and applies consistently across both convolutional networks and transformers is itself an engineering challenge; dependency-graph pruning was proposed precisely to make structured pruning generalize across
architectures \cite{fang2023depgraph}. As a result, compressing a new model well
still depends heavily on human expertise and trial and error.

\subsubsection{Uniform Compression Ignores Layer Sensitivity}
Many existing methods apply pruning and quantization uniformly, using a single
pruning ratio and a single bit-width for the entire model. This ignores the fact
that layers differ widely in how sensitive they are; removing or heavily quantizing one layer may have little effect, while doing the same to another can sharply degrade accuracy \cite{dong2019hawq}, and pruning one part of the model can also change the behavior of the layers that depend on
it \cite{fang2023depgraph,cheng2024survey}. A uniform setting therefore compresses
some layers too aggressively and others too little at any given target. Even methods that support non-uniform pruning ratios or mixed precision often still rely on a human to decide which layers to prune, by how much, and at what bit-width, or on hand-designed sensitivity proxies rather than on measured, empirical sensitivity; and in many cases their algorithms are tied to a particular architecture and cannot easily be extended to others. This burden is made worse by the fact that the right choice is not fixed; the same architecture can behave differently on different datasets, so a configuration that works in one setting does not necessarily carry over to another. Designing an algorithm that can make these per-layer pruning and quantization decisions automatically, and adapt them to the given model and dataset, is therefore difficult.

\subsubsection{Data-Hungry Joint Optimization}
Some approaches, such as GETA~\cite{qu2025automatic}, are data-hungry; because they re-optimize the entire network into a sparse, low-precision subnetwork through end-to-end gradient training, they need a large amount of training data to compress the model while keeping its accuracy high. As we show later, under a limited training budget, GETA fails to recover accuracy, whereas our method remains data-efficient. This makes such methods difficult to use in settings where data or compute is limited.

\subsubsection{Profiling Disconnected from Decisions}

Compressing a model well requires knowing where its redundant parts are for a
specific dataset. This means using data to measure how the model actually reacts to compression; building a per-layer sensitivity map, and understanding how parameters and computation are distributed across the architecture~\cite{lorentz2022profiling}. Profiling tools can provide exactly this kind of information, but in practice they are typically used for manual inspection rather than to drive the compression process automatically~\cite{hu2022dpro}. As a result, the measurement of where a model spends its cost and where it is redundant stays disconnected from the decisions
about what to prune and quantize, even though understanding how compression affects the model should be one of the main inputs to those decisions.

\subsection{Proposed Approach and Contributions}

To address these problems, we propose APQF, an agentic, profiling-guided framework
for model optimization that combines structured pruning, mixed-precision
quantization, and fine-tuning in a single automated pipeline. APQF is built from a
set of cooperating agents that profile a target model, decide how to compress it,
and recover its accuracy, so that the core compression decisions are made
automatically and grounded in measured evidence rather than manual heuristics. To
our knowledge, APQF is the first framework to combine LLM-guided,
profiling-grounded decisions with a fully training-aware pipeline that performs
sequential structured pruning and mixed-precision quantization-aware training with
recovery, across both convolutional and transformer architectures.The main
contributions of this thesis are as follows:

\begin{itemize}
  \item \textbf{Profiling agent.} A profiling component that produces a per-layer
  sensitivity map and a compact architecture brief, showing how parameters and
  computation are distributed, where the cost bottlenecks are, and how the model
  reacts to pruning. This measured evidence grounds all subsequent compression
  decisions.

    \item \textbf{LLM-guided pruning and recovery.} 
  Multi-stage structured pruning in which
the LLM assigns a per-layer pruning ratio from the profiling data, followed by adaptive
recovery in which the LLM selects the fine-tuning method (PEFT or full) and its
hyperparameters, all in an automated process.

  \item \textbf{LLM-guided mixed-precision QAT with distillation.} Quantization-aware
  training in which the LLM assigns a per-layer bit-width and the quantized model is
  trained under a knowledge-distillation objective against a full-precision teacher,
  with an additional optional distillation stage that further recovers accuracy after
  quantization. The strategy and its hyperparameters are chosen by the LLM with
  minimal user intervention.

  \item \textbf{Architecture-agnostic pipeline.} A single framework that operates on
  both convolutional and transformer architectures (ResNet, VGG, ViT, DeiT, and Swin)
  without architecture-specific rules.

    \item \textbf{Data-efficient compression.} APQF prunes a pretrained model from
  measured sensitivity and recovers it with parameter-efficient or full fine-tuning
  together with knowledge distillation. In several of our experiments, this reaches
  the accuracy of competing methods with less training data or fewer training
  epochs, because it does not re-optimize the entire network from scratch.

      \item \textbf{LLM-agnostic and future-proof design.} Because APQF makes its
  decisions through an LLM accessed via a standard API, it is not tied to any single
  model; it works with a range of planners, from inexpensive open models to premium
  ones, and can directly benefit from stronger future LLMs without any change to the
  framework.

\end{itemize}

\section{Related Work}
The expansion of deep learning has accelerated the adoption of deep neural networks. However, the significant improvement in model accuracy has been accompanied by a substantial increase in model depth, width, and parameter count, resulting in much higher computational complexity, memory size, and energy consumption \cite{zhu2025comprehensive}. These increasing resource requirements have become a key challenge to deploying modern deep neural networks on resource-constrained platforms, including smartwatches, IoT nodes, and intelligent sensors \cite{ngo2025edge}. Consequently, considerable research efforts have focused on developing efficient model compression techniques that reduce computational cost while maintaining predictive performance. Moreover, the popularity of Large Language Models (LLMs) in recent years has led to a growing interest in compressing neural networks for devices with flexible hardware requirements \cite{frantar2023sparsegpt}. Among these techniques, pruning, quantization, and knowledge distillation are the most widely accepted approaches for reducing inference cost and model size, while parameter-efficient fine-tuning (PEFT) has become a standard tool for adapting and recovering models at low training cost \cite{rafat2023mitigating, malihi2023efficient, argerich2024measuring, xu2026parameter}.

\subsection{Profiling}
Model profiling evaluates a neural network's computational and structural characteristics by extracting execution information at the model or layer level. Details such as computational cost, memory consumption, latency, and resource utilization help identify performance bottlenecks and deployment constraints \cite{hundt2026xprof,Li_2020}. Unlike theoretical metrics such as FLOPs (Floating Point Operations) \cite{hunger2005floating}, profiling reflects a model's actual runtime behavior on the target hardware and is commonly used to guide compression by identifying expensive or redundant components to prune or quantize \cite{lorentz2022profiling}. Tools such as PyTorch Profiler \cite{pytorch_profiler}, TensorFlow Profiler \cite{tensorflow_profiler}, NVIDIA nvprof \cite{nvidia_profiler}, and CUPTI \cite{nvidia_cupti} expose fine-grained runtime traces, but these are typically used for manual inspection rather than to drive automated compression \cite{hu2022dpro}.

In APQF, profiling directly guides compression. The profiling agent uses ptflops \cite{ptflops} to measure the per-layer distribution of parameters and multiply--accumulate operations (MACs) and the PyTorch Profiler \cite{pytorch_profiler} to record execution time, and it performs an empirical sensitivity analysis by pruning each structural group and measuring the resulting accuracy change. This information is aggregated into a compact architecture summary that grounds the LLM's pruning and quantization decisions in measured evidence.

\subsection{Pruning}
Pruning is a model compression technique that removes redundant weights or structures from a neural network to lower its size and computation. It works by identifying redundant weights, channels, or layers in the model and removing them with little or no effect on performance \cite{li2023model}. Pruning can be categorized into Structured \cite{filters2016pruning}, Unstructured~\cite{han2015learning}, and Semi-Structured Pruning \cite{cheng2024survey}. These three techniques differ mostly in the level at which parameters are removed and in
their deployment characteristics. Unstructured pruning removes individual weights and typically produces the largest compression ratio \cite{he2023structured}. Nevertheless, its irregular sparsity frequently necessitates the use of specialized sparse compute libraries or hardware to achieve inference speedups \cite{cheng2024survey}. Semi-structured pruning removes weights using predetermined sparsity patterns in order to attain both structural regularity and high model accuracy. Representative patterns, such as 2:4 sparsity, allow efficient execution on supported hardware while reducing the accuracy loss often observed at high pruning ratios \cite{frantar2023sparsegpt,cheng2024survey}. Structured pruning, on the other hand, removes entire network components, producing a regular architecture that can be executed quickly on standard hardware without the need for specialized sparse compute support. Structured pruning is a popular model compression technique for resource-constrained applications due to its efficient deployment and practical acceleration \cite{cheng2024survey}. APQF  adopts structured pruning. Specifically, it builds on DepGraph \cite{fang2023depgraph}, a dependency-graph-based method that groups coupled channels and prunes them consistently across layers so that the pruned model remains structurally valid.

\subsection{Quantization}
Quantization reduces the numerical precision of a model to lower its storage cost and enable efficient inference on integer hardware \cite{zhou2017incremental}. It represents model weights and activations using low-precision values. Quantization methods are generally categorized as post-training quantization (PTQ) or quantization-aware training (QAT) \cite{zhao2023post}. PTQ quantizes a pretrained model without retraining, making it computationally efficient. However, because the parameters are not adapted to the quantized representation, PTQ can suffer from substantial accuracy degradation, particularly at low bit-widths \cite{liu2023pd}. In contrast, QAT simulates quantization during training, allowing the model to adapt to low-precision weights and activations. QAT generally preserves accuracy better than PTQ under aggressive quantization, although it requires additional training \cite{chen2025efficientqat}.

Early low-precision methods include ternary and binary weight quantization, such as TWN and LR-Net \cite{li2016ternary,shayer2017learning}. RQ and WAGE extend low-precision optimization to weights and activations \cite{louizos2018relaxed,wu2018training}, while DQ learns mixed-precision configurations through differentiable optimization \cite{uhlich2019mixed}. Other approaches, including DJPQ, Bayesian Bits, and GETA, combine pruning with mixed-precision quantization in unified compression procedures \cite{wang2020differentiable,van2020bayesian,qu2025automatic}.

Because APQF targets high accuracy under aggressive compression, it combines structured pruning with mixed-precision QAT. After pruning, the framework assigns a bit-width to each surviving layer and simulates low-precision execution using Brevitas \cite{brevitas}.

\subsection{Knowledge Distillation}
Knowledge distillation (KD) transfers knowledge from a large, complex teacher model to a smaller, lightweight student model while maintaining performance \cite{xu2026deep}. This is commonly achieved by training the student with a combined loss that blends hard-target supervision from the ground-truth labels with the softened output distribution of the teacher \cite{yang2025feature}. Learning from the teacher's soft predictions, rather than from labels alone, transfers richer information about class relationships and allows the student to approach the teacher's accuracy at a fraction of its size \cite{moslemi2024survey}. As a result, KD is widely used to deploy efficient models on resource-constrained devices without significant accuracy loss. APQF employs response-based knowledge distillation following Hinton et al. \cite{hinton2015distilling}. A frozen full-precision teacher guides the quantized student through a weighted combination of temperature-scaled KL divergence and hard-label cross-entropy.

\subsection{Parameter-Efficient Fine-Tuning (PEFT)}
Foundation models now dominate a wide range of domains, such as language-specific and multimodal tasks. To achieve task-specific performance in real-world applications, these models are typically fine-tuned on unseen downstream datasets \cite{zhang2025parameter}. Parameter-Efficient Fine-Tuning (PEFT) is a cost-effective fine-tuning strategy that reduces the number of updated parameters and the computational complexity while still achieving suitable downstream task performance \cite{wang2025parameter}. Compared to full fine-tuning, which updates all model parameters, PEFT modifies only a small number of parameters while keeping the majority of the pretrained model frozen. Existing PEFT methods can be broadly categorized into adapter-based, prompt-based, and selective methods \cite{yuan2025efficientllm}. Among these, low-rank adaptation methods have been widely adopted due to their simplicity, parameter efficiency, and strong performance \cite{hu2022lora}.

Several low-rank variants have since been proposed, including LoRA \cite{hu2022lora}, DoRA \cite{liu2024dora}, PiSSA \cite{meng2024pissa}, rsLoRA \cite{kalajdzievski2023rank}, and LoRA+ \cite{hayou2024lora+}, which differ in how they parameterize, initialize, scale, or optimize the low-rank adapters. APQF uses these methods for post-pruning accuracy recovery, and we describe each one in detail in the Methodology section.

\subsection{Agentic Reasoning Frameworks}
Agentic reasoning is the component of an AI agent that handles decision making and allows the agent to perform tasks autonomously, using conditional logic or heuristics and relying on perception and memory to pursue goals and optimize for a successful outcome. In this setting, large language models (LLMs) are increasingly used not only as generative tools but as autonomous agents that carry out multi-step reasoning by interacting with tools or environments \cite{wu2025agentic}. Prompting strategies such as Chain-of-Thought \cite{wei2022chain, madaan2022text} show that LLMs can construct their own step-by-step reasoning process, while ReAct \cite{yao2023react} and STaR \cite{zelikman2022star} demonstrate that they can generate intermediate reasoning steps that improve both interpretability and accuracy. These abilities extend to tasks requiring sequential planning, tool use, and API invocation, allowing LLMs to interact with and control external systems \cite{ahn2022can, lin2023swiftsage, park2023generative, schick2023toolformer}. Unlike conventional LLMs that produce a response in a single forward pass, agentic frameworks maintain a dynamic context throughout the reasoning process and adapt to new tasks through in-context learning without retraining \cite{zhao2025llm}. Such frameworks are commonly categorized into single-agent, tool-augmented, and multi-agent approaches, and have shown strong performance on complex reasoning problems, establishing agentic reasoning as a practical approach for autonomous and organized decision making in modern foundation models.

Building such agentic systems in practice requires a convenient way to access and switch between different LLMs. OpenRouter is a unified gateway service that addresses this challenge by routing requests to different models, allowing multiple LLMs to be integrated into a single application through one interface \cite{roumeliotis2023chatgpt}. It provides access to a broad selection of models from multiple platforms \cite{morillo2026scalable} through a common API, exposing a broad selection of commercial and open-source models under one interface. In this work, we use OpenRouter to access the LLMs used by the different agents in APQF, which allows us to run the framework with a range of LLM planners through a single, consistent interface.

\subsection{Agentic and Joint Compression Methods}

The most closely related work either guides compression with LLM-based agentic reasoning or jointly optimizes pruning and quantization. Our prior work, ProfilingAgent \cite{jafari2025profilingagent}, introduced a profiling-guided multi-agent LLM system that automates structured pruning and post-training dynamic quantization by reasoning over static metrics such as MACs and parameter counts and dynamic signals such as latency and memory to design architecture-specific strategies. In a similar spirit, Kodathala and Vunnam \cite{kodathala2026llms} use a foundation model as an adaptive pruning agent that selects which layers to prune from weight--activation and gradient sensitivity profiles, using self-reflection and checkpoint rollback to preserve knowledge in large language models.

A second line of work jointly optimizes pruning and quantization without an LLM. GETA \cite{qu2025automatic} performs automatic joint structured pruning and quantization-aware training on arbitrary networks through a quantization-aware dependency graph and a partially projected gradient method that enforces layer-wise bit constraints. DJPQ \cite{wang2020differentiable} formulates joint pruning and mixed-precision quantization as a single differentiable objective, combining variational information-bottleneck structured pruning with learned bit-widths to reduce bit-operations (BOPs). Bayesian Bits \cite{van2020bayesian} unifies quantization and pruning through a gradient-based decomposition that doubles the bit-width in residual steps, where an additional 0-bit option treats pruning as the extreme case of quantization.

APQF builds on this prior work but differs by combining LLM-guided, profiling-grounded decisions with a fully training-aware pipeline, applying multi-stage structured pruning, adaptive PEFT-based recovery, and mixed-precision quantization-aware training with knowledge distillation across both convolutional and transformer architectures.

\section{Methods And Procedures}
APQF is a profiling-guided model compression through structured pruning followed by quantization. The framework is organized around five main agents: the ProfilingAgent, PrunerAgent, FineTuningAgent, QuantAgent, and EvaluationAgent. Together, these agents form an end-to-end pipeline that profiles a target vision model, applies LLM-guided multi-stage structured pruning, performs adaptive fine-tuning for post-pruning accuracy recovery, applies mixed-precision quantization-aware training, performs post-quantization recovery using knowledge distillation, and finally evaluates the compressed model. We apply APQF to both convolutional and transformer-based image classification models, including ResNet \cite{he2016deep}, ViT \cite{dosovitskiy2020image}, DeiT \cite{touvron2021training}, and Swin \cite{liu2021swin}, on the ImageNet \cite{deng2009imagenet} and CIFAR-10 \cite{krizhevsky2009learning} datasets.

\begin{figure}[!htbp] \centering
\includegraphics[width=0.5\textwidth]{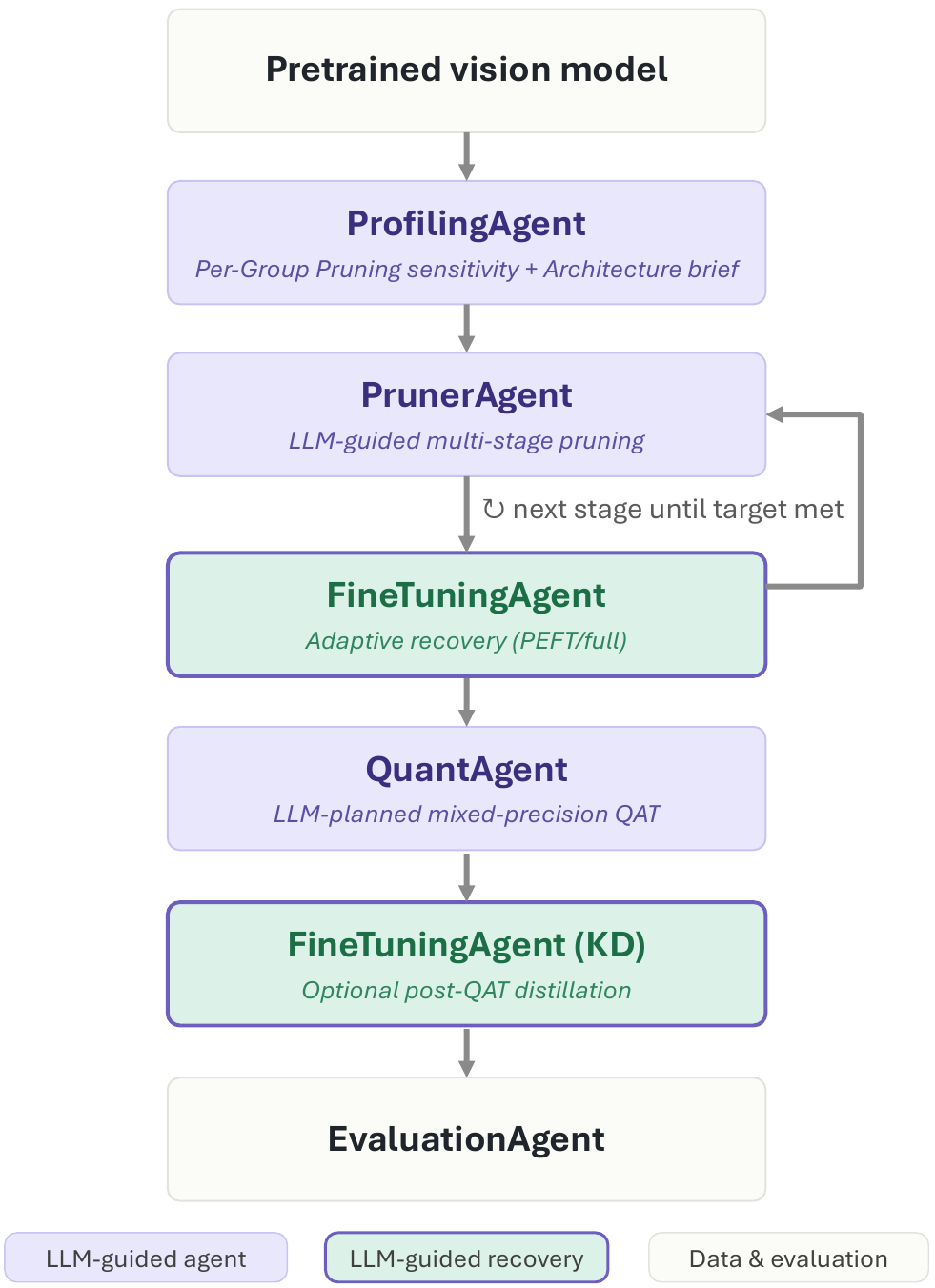}
\caption{Overall workflow of the APQF framework}
\label{workflow}
\end{figure}

As shown in Figure \ref{workflow}, APQF starts with a pretrained vision model and first sends it to the ProfilingAgent. The ProfilingAgent analyzes the model structure, measures pruning sensitivity, and produces an architecture brief. This information is then used by the PrunerAgent to perform LLM-guided multi-stage structured pruning toward a target compression level. After each stage, the FineTuningAgent adaptively recovers the lost accuracy, and the pipeline returns to the PrunerAgent for the next stage until the target parameter reduction is met. Once the pruning target is reached, the QuantAgent applies LLM-planned mixed-precision quantization-aware training, followed by an optional post-QAT knowledge distillation recovery stage. Finally, the EvaluationAgent measures the accuracy and compression performance of the final model.

This section describes each component of the APQF framework in detail. It explains the role of each agent, how information flows between agents, and how the overall pipeline combines profiling, pruning, recovery, quantization, and evaluation to study the accuracy-compression trade-off of vision models.

\subsection{Profiling Agent}
The ProfilingAgent is the first analysis stage in APQF. The purpose of the ProfilingAgent is to build a detailed understanding of the target model before any compression is applied. Modern vision models contain a large number of layers and modules whose contributions to accuracy are highly uneven.  Some are redundant and can be pruned with little effect, while others are sensitive and largely determine the model's accuracy. A central design goal of APQF is to be architecture-agnostic, so that the framework can operate on model families ranging from convolutional networks (CNNs) to vision transformers (ViTs). For this reason, it analyzes the model at the level of modules, layers, blocks, and dependency groups rather than relying on a fixed architecture-specific pruning rule.

\begin{figure}[!htbp] \centering
\includegraphics[width=0.7\textwidth]{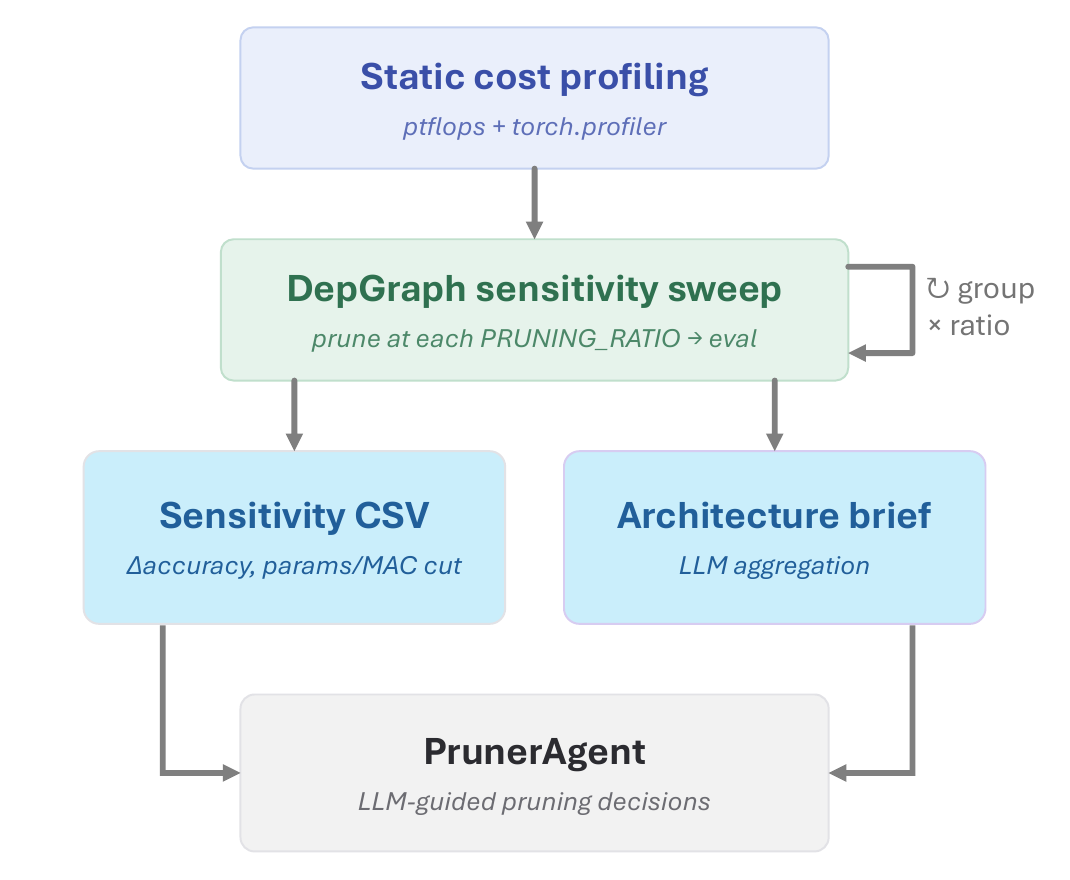}
\caption{ProfilingAgent workflow producing the sensitivity CSV and architecture brief}
\label{Profiling}
\end{figure}

As shown in Figure \ref{Profiling}, the ProfilingAgent performs two complementary forms of profiling. The first is static cost profiling, which produces a map of the model's resource usage. Using \texttt{ptflops} \cite{ptflops}, the agent measures how multiply–accumulate operations (MACs) and parameters are distributed across the modules, blocks, and layers of the network, and using \texttt{torch.profiler} \cite{pytorch_profiler} it records the CPU and GPU execution time of the different parts of the model. Because this raw profiling output is verbose and would overwhelm the reasoning of the downstream agents, it is passed to an LLM that aggregates it into a compact architecture brief summarizing the model's structure and its cost distribution.

The second and most important form of profiling is the DepGraph sensitivity sweep, which empirically measures how prunable each structural group is. Given a set of pruning ratios specified by the user, the agent uses dependency-graph (DepGraph) structured pruning \cite{fang2023depgraph} to prune each architecture-aware group at each ratio, evaluates the resulting model, and records the corresponding accuracy drop together with the achieved parameter and MAC reduction. Repeating this over all groups and ratios yields a per-group accuracy–cost sensitivity table, which is saved as a CSV file.

Together, the sensitivity CSV and the architecture brief constitute the output of the ProfilingAgent. These two artifacts are consumed by the downstream pruning and quantization agents, which rely on them to make informed, model-specific compression decisions.

\subsection{Pruning Agent}
\label{subsec:PrunnerAgent}
The PrunerAgent is the first compression agent in APQF. Its role is to reduce the number of model parameters through LLM-guided structured pruning. Before pruning begins, the user specifies the target parameter-reduction percentage and the maximum number of pruning stages. Rather than pruning the model to the target in a single step, the PrunerAgent approaches the target gradually through a sequence of LLM-guided pruning stages, interleaved with adaptive accuracy recovery.

\begin{figure}[!htbp] \centering
\includegraphics[width=0.7\textwidth]{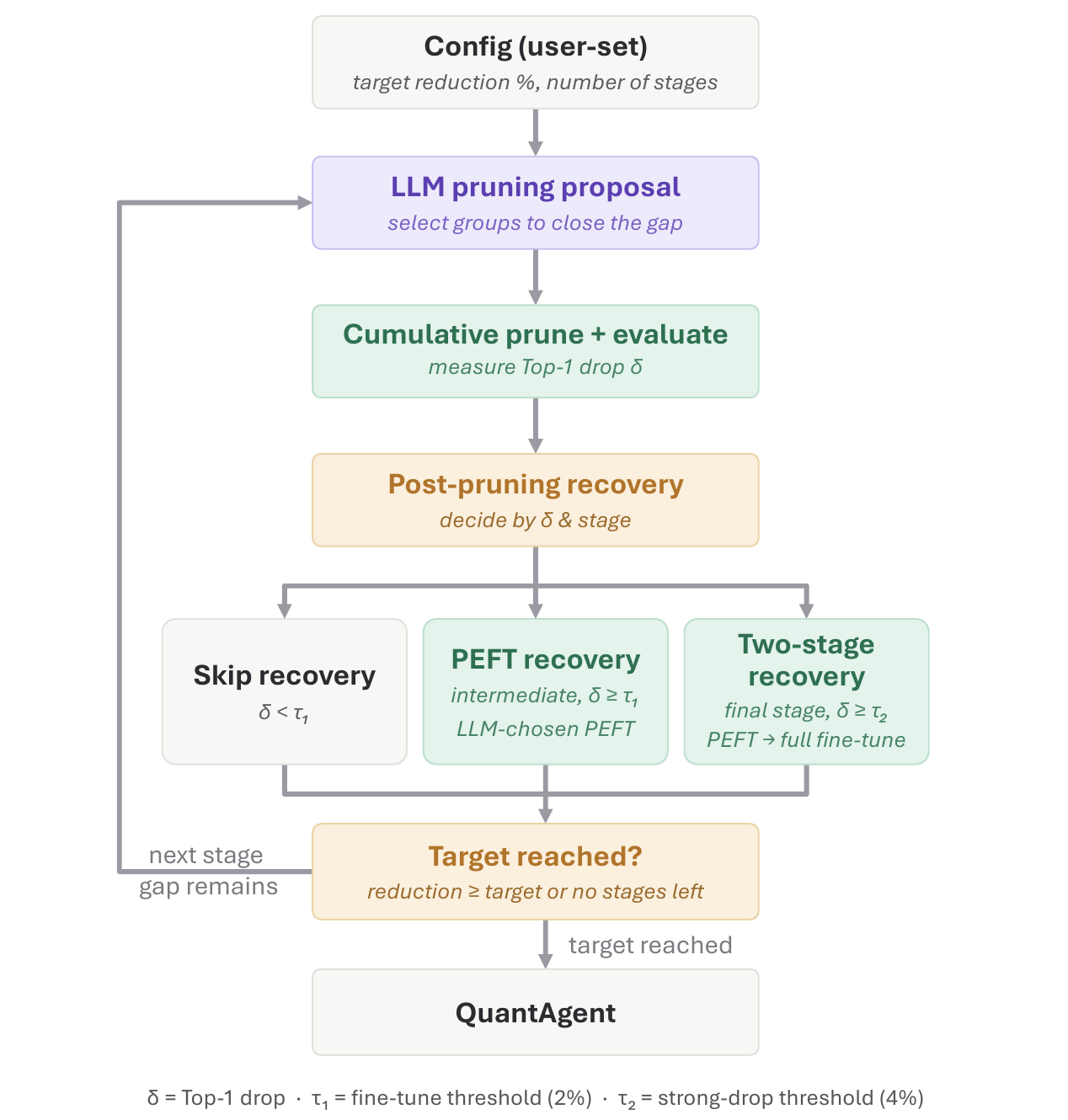}
\caption{Multi-stage target-seeking pruning with adaptive fine-tuning}
\label{Pruner}
\end{figure}

As shown in Figure \ref{Pruner}, pruning is performed as a multi-stage, target-seeking process. The PrunerAgent is guided by the profiling data produced once by the ProfilingAgent including the per-group sensitivity table and the architecture brief, which it reuses across all stages. At the start of each stage, it issues a prompt asking the LLM to propose a pruning strategy, combining this profiling data with the user's compression target and the current state of the process, the parameter reduction achieved so far, the remaining gap to the target, the groups selected and recovery decisions made in previous stages. APQF builds on DepGraph structured pruning \cite{fang2023depgraph}, which groups coupled channels so they can be pruned consistently across layers. Unlike standard DepGraph, which applies a single user-set pruning ratio, APQF lets the LLM assign a per-group (per-layer) pruning ratio, making the structured pruning LLM-guided rather than uniform. The LLM proposes a pruning strategy by selecting candidate dependency groups that are expected to remove redundant parameters while causing minimal accuracy degradation. Before the strategy is applied, APQF validates the proposed groups against the measured profiling data to ensure that the selected groups, pruning ratios, and layer names are valid before pruning is executed.

The PrunerAgent operates in close interaction with the FineTuningAgent, as illustrated in Figure \ref{Pruner}. After a stage is pruned, the model is evaluated and its Top-1 accuracy drop~$\delta$ is measured. The framework then adaptively decides how to recover the lost accuracy based on user-defined thresholds. When the accuracy drop is small ($\delta < \tau_1$), recovery is skipped, since fine-tuning would provide little benefit and may even degrade the model through overfitting. When the drop is moderate ($\delta \geq \tau_1$) on an intermediate stage, the model is recovered using parameter-efficient fine-tuning (PEFT) before proceeding. On the final stage, when the accuracy drop is large ($\delta \geq \tau_2$), the framework applies a stronger two-stage recovery consisting of a PEFT phase followed by full fine-tuning. The mechanics of these recovery procedures are described in Section \ref{subsec:finetuning}.

A key design choice is how pruning is applied across stages. Instead of applying
each new pruning stage directly on top of the previously pruned model, APQF
accumulates the selected dependency groups across stages and replays the
cumulative pruning strategy from the original model. This avoids a common issue
in dependency-graph pruning, where pruning an already-pruned checkpoint may
produce little additional parameter reduction and stall before the target is
reached. In this design, the stages accumulate structural pruning
\emph{decisions}, while intermediate evaluation and recovery provide feedback
for the next LLM-guided decision. The final post-pruning checkpoint is selected
from the last cumulative pruning stage, optionally after adaptive fine-tuning
when recovery is triggered and improves accuracy.

Intermediate fine-tuning is included to make the pruning loop recovery-aware.
Rather than judging each pruning stage only by the immediate post-pruning
accuracy drop, APQF evaluates how much of the degradation can be recovered. This
provides the next pruning stage with a more realistic estimate of the remaining
model capacity and prevents the LLM planner from becoming overly conservative
after temporary accuracy loss. Thus, multi-stage fine-tuning serves as an
accuracy-feedback mechanism for target-seeking pruning, while the cumulative
pruning strategy continues to control structural parameter reduction.

This process repeats stage by stage including pruning, evaluating, and adaptively recovering until the measured cumulative parameter reduction reaches the user's target (within a tolerance) or the maximum number of stages is exhausted. The complete procedure is summarized in Algorithm \ref{alg:prunner}.

\begin{algorithm}[!htbp]
\small
\setlength{\interspacetitleruled}{0pt}
\setlength{\algomargin}{1.2em}
\DontPrintSemicolon
\caption{Multi-stage target-seeking pruning with adaptive recovery}
\label{alg:prunner}
\KwIn{model $M_0$; profiling data (sensitivity table, architecture brief);
      target reduction $\rho^\star$; stages $N$; tolerance $\epsilon$; thresholds $\tau_1 < \tau_2$}
\KwOut{final post-pruning checkpoint $M$; selected groups $G$}
\BlankLine
$M \leftarrow M_0$;\quad $G \leftarrow \emptyset$ \tcp*{selected groups}
\For{$k = 1$ \KwTo $N$}{
  $\rho \leftarrow$ parameter reduction of $M$ \tcp*{relative to $M_0$}
  $G_k \leftarrow \textsc{LlmPropose}(\rho^\star, \rho, \rho^\star-\rho, \text{history})$ \tcp*{target, current, gap, history}
  \If{$G_k$ invalid}{
    $G_k \leftarrow \textsc{Repair}(G_k)$ \tcp*{one repair retry}
    \lIf{$G_k$ invalid}{\Return failure}
  }
  $G \leftarrow G \cup G_k$\;
  $M \leftarrow \textsc{Prune}(M_0, G)$ \tcp*{cumulative replay from original}
  $\rho \leftarrow$ parameter reduction of $M$;\quad $\delta \leftarrow$ Top-1 drop of $M$\;
  $\mathit{reached} \leftarrow (\rho \geq \rho^\star - \epsilon)$;\quad
  $\mathit{last} \leftarrow \mathit{reached} \textbf{ or } (k = N)$ \tcp*{last pruning stage?}
  \uIf{$\delta < \tau_1$}{
    keep pruned $M$ \tcp*{recovery not needed}
  }
  \Else{
    \uIf{$\mathit{last}$ \textbf{and} $\delta \geq \tau_2$}{
      $M_{\mathrm{ft}} \leftarrow \textsc{TwoStage}(M)$ \tcp*{PEFT, then full fine-tune}
    }
    \Else{
      $M_{\mathrm{ft}} \leftarrow \textsc{Peft}(M)$ \tcp*{light recovery}
    }
    $M \leftarrow$ better of pruned $M$ and $M_{\mathrm{ft}}$ \tcp*{keep-better}
  }
  \lIf{$\mathit{reached}$}{\Return $M$}
}
\Return $M$\;
\BlankLine
\footnotesize
Here $M_0$ is the original model, $M$ the current checkpoint, $G$ the cumulative set of
selected pruning groups and $G_k$ the groups added at stage~$k$, $\delta$ the Top-1
accuracy drop, and $M_{\mathrm{ft}}$ the fine-tuned candidate compared against the pruned
model in the keep-better step.
\end{algorithm}

\subsection{Quantization Agent}
\label{subsec:Quantization}
The QuantAgent applies LLM-guided mixed-precision quantization-aware training (QAT) to the pruned model produced by the PrunerAgent. To decide how aggressively each layer can be quantized, it draws on the same profiling evidence used earlier, including the ProfilingAgent architecture brief and the structured-pruning sensitivity table. This information comes with information specific to the pruned model, such as the PrunerAgent results from the preceding pruning stage and the actual list of quantizable modules (the surviving convolutional and linear layers) extracted from the pruned checkpoint. The agent also verifies that the checkpoint's parameter count matches the pruning results, ensuring that the quantization strategy is generated for the correct pruned model.

\begin{figure}[!htbp] \centering
\includegraphics[width=0.7\textwidth]{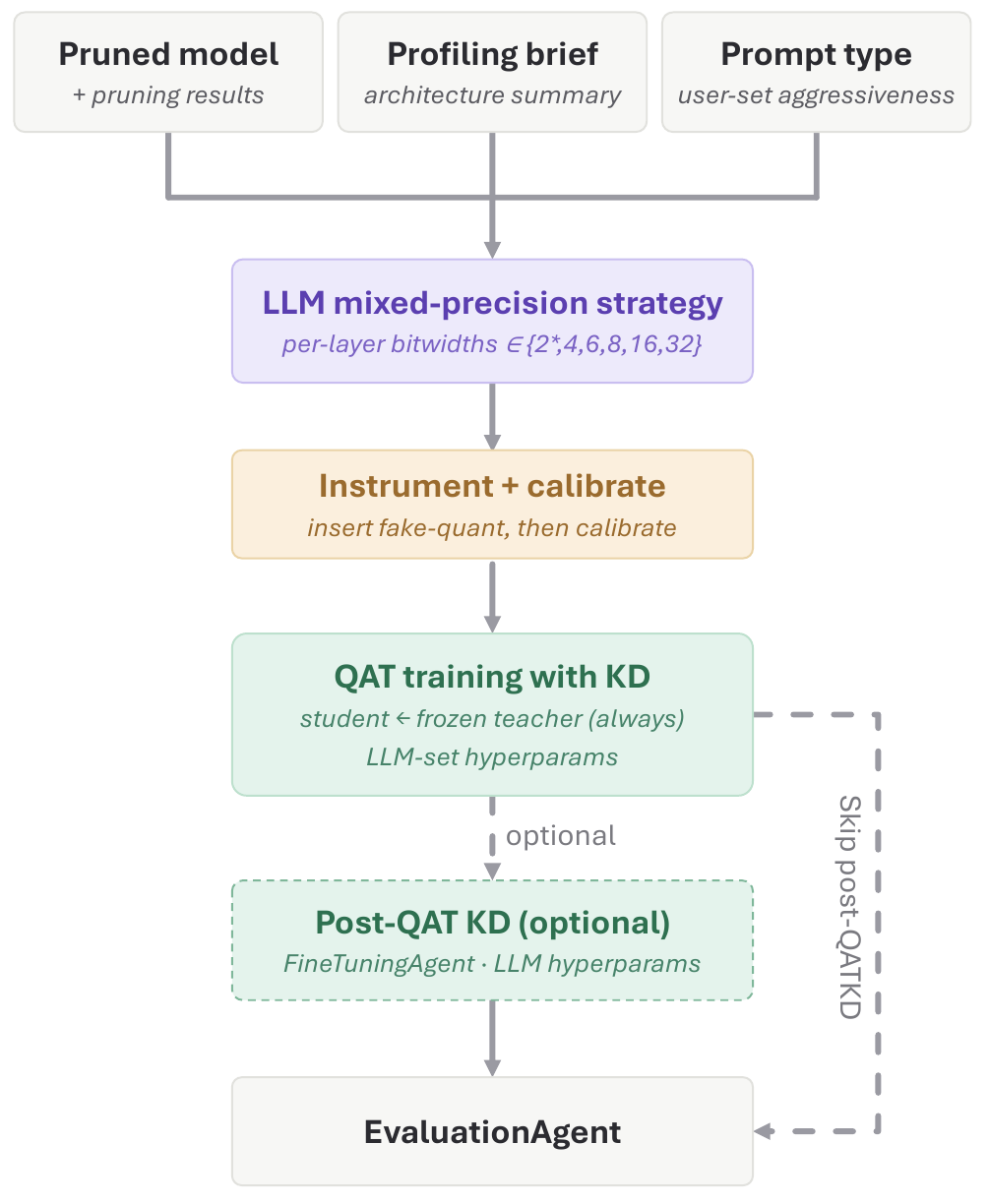}
\caption{QuantAgent workflow for LLM-guided mixed-precision quantization}
\label{Quant}
\end{figure}

As shown in Figure \ref{Quant}, the LLM proposes a mixed-precision strategy that assigns a bit-width to each quantizable layer, chosen from ${2, 4, 6, 8, 16, 32}$, where the lowest 2-bit setting is reserved for specific architectures (e.g., VGG). The proposal is conditioned on the profiling evidence and on a user-selected prompt type, including \emph{conservative}, \emph{balanced}, or \emph{aggressive},  which controls how strongly the strategy favors lower bit-widths. Before training, APQF validates the proposed strategy against the real surviving modules, rejecting invalid layer names or bit-widths so that quantization is only applied to modules that actually exist in the pruned model.

The validated strategy is then implemented by instrumenting the model with
simulated quantization modules from Brevitas~\cite{brevitas} at the assigned
bit-widths and calibrating them on a small data sample. Because this is quantization-aware training, the quantized model is trained rather than quantized in a single shot. The quantized student is optimized via knowledge distillation
(KD)~\cite{hinton2015distilling} from a frozen full-precision teacher.  Let $z_s$ and $z_t$ be the student and teacher logits, $\sigma(\cdot)$ the softmax,
$T$ the distillation temperature, and $y$ the ground-truth label. Define the
temperature-softened output distributions $q_s = \sigma(z_s/T)$ and
$q_t = \sigma(z_t/T)$. The quantized student minimizes
\begin{equation}
\mathcal{L}_{\mathrm{KD}} = T^{2}\,\mathrm{KL}\!\left(q_t \,\|\, q_s\right),
\qquad
\mathcal{L}_{\mathrm{CE}} = \mathrm{CE}(z_s, y),
\label{eq:kd}
\end{equation}
where $\mathrm{KL}(\cdot\,\|\,\cdot)$ denotes the Kullback--Leibler divergence~\cite{kullback1951information} between
the softened teacher and student distributions, $\mathrm{CE}(\cdot,\cdot)$ is the
cross-entropy against the ground-truth label, and the factor $T^{2}$ rescales the
distillation gradient so that it remains comparable in magnitude to the
cross-entropy term. The two losses are combined through a KD weight
$\alpha \in [0,1]$:
\begin{equation}
\mathcal{L} = \alpha\,\mathcal{L}_{\mathrm{KD}} + (1-\alpha)\,\mathcal{L}_{\mathrm{CE}},
\label{eq:qat_loss}
\end{equation}
where $\mathcal{L}_{\mathrm{KD}}$ is the soft-target distillation loss that encourages the student to reproduce the teacher's full class distribution, and
$\mathcal{L}_{\mathrm{CE}}$ is the standard hard-label classification loss.

The LLM selects the QAT training hyperparameters, such as the learning rate, number of epochs, warmup steps, weight decay, batch size, and number of calibration samples, to maximize the accuracy recovery under the chosen bit-width assignment, while the distillation temperature and KD weight can be selected by the user.

Finally, APQF supports an optional post-QAT KD stage. Whereas the KD inside QAT guides the model \emph{while it learns} its low-bit representation and is an inherent part of quantization, the post-QAT KD is a separate recovery pass applied to the already-quantized model after QAT has finished. It uses the same distillation objective but serves only to recover additional accuracy when QAT has degraded the model; it is optional and is kept only if it improves accuracy. This stage is carried out by the FineTuningAgent and is described in Section~\ref{subsec:finetuning}.

\subsection{FineTuning Agent}
\label{subsec:finetuning}
The FineTuningAgent is a shared recovery component that is invoked by two other agents in APQF. It is called by the PrunerAgent to restore accuracy after structured pruning (Section~\ref{subsec:PrunnerAgent}), and by the QuantAgent to perform the optional post-quantization recovery (Section~\ref{subsec:Quantization}). In both roles, the recovery method and its hyperparameters are chosen by an LLM planner, which adapts the recovery to the specific model and to the amount of degradation observed, and every proposed plan is validated before it is executed.

\subsubsection{Post-pruning recovery}

For post-pruning recovery, the FineTuningAgent supports full fine-tuning, several parameter-efficient fine-tuning (PEFT) methods, and a two-stage recovery mode. In full fine-tuning, all model parameters are updated using a small learning rate. This provides stronger recovery capacity but is more expensive and can overfit when the recovery set is limited. In PEFT recovery,
only a small number of additional adapter parameters are trained while most of the pruned backbone remains frozen. This makes PEFT suitable for intermediate pruning stages, where the goal is to recover enough accuracy to guide the next pruning decision without fully retraining the model.APQF implements five PEFT variants:

\begin{itemize}
\item \textbf{LoRA} (Low-Rank Adaptation) injects a trainable low-rank update $\Delta W = BA$ into each target layer while freezing the original weights, so that only the small matrices $A$ and $B$ (of rank $r$) are trained~\cite{hu2022lora}.
\item \textbf{DoRA} (Weight-Decomposed Low-Rank Adaptation) decomposes each weight into a magnitude and a direction and applies a low-rank update to the directional component, improving stability and expressiveness over standard LoRA~\cite{liu2024dora}.
\item \textbf{rsLoRA} (rank-stabilized LoRA) rescales the low-rank update so that training remains stable at higher adapter ranks, allowing larger ranks without degrading convergence~\cite{kalajdzievski2023rank}.
\item \textbf{PiSSA} starts the low-rank update from the most significant directions of the pretrained weight matrix instead of random values, which improves the starting point. It applies to linear layers only, so it is not used for convolutional (CNN) recovery~\cite{meng2024pissa}.
\item \textbf{LoRA+} assigns different learning rates to the two low-rank matrices, accelerating convergence relative to standard LoRA~\cite{hayou2024lora+}.
\end{itemize}

The LLM planner selects which method to use for a given model, together with its hyperparameters, such as the adapter rank, the scaling factor, the learning rate, and the number of epochs. The proposed plan is validated before execution; for instance, PiSSA is rejected for convolution-only models.

For the strongest recovery, which is used at the final pruning stage when the accuracy drop is large, the FineTuningAgent applies a two-stage procedure. In the first stage, a PEFT method (for example, rsLoRA) is trained as a warm-up, and its adapters are then merged back into the weights; in the second stage, all parameters are unfrozen and fine-tuned with a small learning rate. This combines the efficiency and stability of PEFT with the higher capacity of full fine-tuning.

\subsubsection{Post-quantization knowledge distillation}
The FineTuningAgent also implements the optional post-QAT recovery invoked by the QuantAgent. In this role, the already-quantized model is treated as the student, and a full-precision model serves as the teacher. The teacher is configurable; it can be either the pruned model or the original uncompressed model, as selected by the user (a specific checkpoint may also be provided). The student is fine-tuned with a small learning rate under the same combined distillation objective used during QAT (Eq.~\ref{eq:qat_loss}), minimizing the soft-target KD loss together with the hard-label cross-entropy loss. Here, the LLM planner selects recovery hyperparameters, including the number of epochs, learning rate, weight decay, warmup steps, distillation temperature, KD weight, and the number of calibration batches. Because this recovery pass may not always help, the resulting checkpoint is retained only if it improves accuracy over the quantized model.

\subsection{Evaluation Agent}
The EvaluationAgent is a deterministic component used to measure the accuracy and efficiency of models produced at different stages of APQF. Unlike the ProfilingAgent, PrunerAgent, QuantAgent, and the fine-tuning planners, the EvaluationAgent does not use an LLM. Instead, it loads the selected checkpoint and evaluates it on a fixed validation manifest, which is a set of validation images drawn once with a fixed random seed and balanced across classes, so that the baseline, pruned, fine-tuned, and quantized models are all compared on the same examples.

For each evaluated model, the EvaluationAgent reports Top-1 accuracy, Top-5 accuracy,
parameter count, and the relative bit-operations (Rel.\ BOPs). Bit-operations weight
each layer's multiply--accumulate count by the precision at which that layer executes,
so that the gains from pruning and from quantization are captured in a single number.
For a network with layers $l = 1,\dots,L$, where layer $l$ performs $M_l$
multiply--accumulate operations and its weights and activations are represented with
$b_w^{(l)}$ and $b_a^{(l)}$ bits,
\begin{equation}
\mathrm{BOPs} \;=\; \sum_{l=1}^{L} M_l \, b_w^{(l)} \, b_a^{(l)}.
\label{eq:bops}
\end{equation}
Pruning reduces $M_l$ by removing channels, while quantization reduces $b_w^{(l)}$ and
$b_a^{(l)}$, so both forms of compression lower the same quantity. Since APQF applies pruning and quantization together, neither the parameter count nor the
bit-width alone reflects the full compression, which makes Rel.\ BOPs a stronger measure
of the compressed model's cost than either metric on its own. The reported metric
normalizes the compressed model against the original full-precision model, in which
every layer runs at $32$ bits,
\begin{equation}
\mathrm{Rel.\ BOPs} \;=\;
\frac{\sum_{l} \tilde{M}_l \, b_w^{(l)} \, b_a^{(l)}}
     {32^{2} \sum_{l} M_l} \times 100\%,
\label{eq:relbops}
\end{equation}
where $\tilde{M}_l$ denotes the multiply--accumulate count of layer $l$ after pruning and
$M_l$ that of the original model. Layers left in full precision keep
$b_w^{(l)} = b_a^{(l)} = 32$. Because the metric is expressed relative to the model's own
full-precision cost, it is architecture-normalized and therefore comparable across
models and across implementations, with lower values indicating a cheaper model. Accuracy and
parameter count are used inside the pipeline, the first to decide whether pruning recovery
is needed and the second to track progress toward the reduction target, while Rel.\ BOPs
is computed at the end to compare the final compressed model against the original
baseline.

\section{Results}
The goal of this section is to assess APQF across different models and datasets. In each case, the objective of compression is to preserve accuracy while making the model cheaper to run by applying pruning and quantization. Because APQF combines structured pruning with quantization, it reduces both the compute and the numerical precision of the model. We summarize these gains primarily through the number of bit operations (BOPs), together with the reduction in model size. Beyond raw compression, we also show that APQF is architecture-agnostic, operating effectively on both convolutional networks (CNNs) and vision transformers (ViTs), and we use a set of ablation studies to demonstrate the importance of the framework's individual novelties, including the role of the LLM in producing better compression decisions.

In the following, we describe the experimental setup, including the models, datasets, hardware, and evaluation metrics; report the main compression and accuracy results alongside comparisons with other methods; and present ablation studies that isolate the contribution of each design choice to the LLM's role in the pipeline.

\subsection{Experimental Setup}
This section describes the datasets, model architectures, implementation settings, evaluation metrics, and comparison methods used to evaluate APQF.

\subsubsection{Datasets and Preprocessing}
We evaluate APQF on the ImageNet-1k and CIFAR-10 image-classification datasets. ImageNet-1K, is made up of 1,281,167 training pictures and 50,000 validation images spread over 1,000 object classes. ImageNet's immense magnitude and variety have made it the recognized benchmark for training and assessing deep learning models in image classification, object identification, and transfer learning~\cite{deng2009imagenet,russakovsky2015imagenet}. CIFAR-10 contains 60,000 color images of size 32 × 32 pixels organized into 10 item classes, with 50,000 for training and 10,000 for testing. Each class comprises 6,000 images from many categories. Because of its small size and balanced class distribution, the CIFAR-10 is widely used for developing and evaluating computer vision algorithms, particularly in research on model compression, pruning, quantization, knowledge distillation, and neural network architecture design \cite{krizhevsky2009learning}.

In APQF, ImageNet and CIFAR-10 are organized in an ImageFolder-style structure. For evaluation, images are preprocessed using each model's associated image processor, which applies the resizing, cropping, and normalization defined by the model's own configuration; the same preprocessing is used for every checkpoint to keep comparisons fair. During fine-tuning and quantization-aware training, we apply standard data augmentation; images are resized to $1.15\times$ the target resolution, randomly cropped with a scale between $0.8$ and $1.0$, horizontally flipped, and color-jittered in brightness, contrast, and saturation, before being converted to tensors and normalized with the model's mean and standard deviation.

\subsection{Model Architectures and Implementation Details}
APQF is designed as a model-agnostic framework whose agents can operate across different vision architectures. To evaluate this capability, we conduct experiments on five representative model families, including the CNN-based ResNet and VGG architectures and the transformer-based ViT, DeiT, and Swin architectures. Where available, APQF loads pretrained or fine-tuned ImageNet-1K and CIFAR-10 checkpoints from the Hugging Face Hub. Hugging Face provides a unified interface for accessing and integrating pretrained computer vision models with PyTorch, enabling consistent model loading across the evaluated architectures~\cite{feki2025empirical,mechalkhexploring}. Locally stored checkpoints are used for models that are not obtained from the Hugging Face Hub.

\textbf{ResNet} has been a breakthrough in deep learning for image classification tasks, and various ResNet variants have been built over the years to improve its performance even more \cite{prajwal2023comparative}. For the purposes of this research, we have explored the ResNet-20 and ResNet-50 model variants; the main difference between them is the number of layers.  ResNet-50 contains 50 layers and about 25.6M parameters. It is a widely used backbone. A smaller configuration, ResNet-20, has 20 layers. These models introduced residual learning to make deep neural networks easier to train \cite{he2016deep}.

\textbf{VGG7} is a lightweight VGG variant that uses stacked \(3 \times 3\) convolutional layers to extract hierarchical image features. Its simple and uniform architecture makes it a useful baseline for evaluating model compression methods~\cite{simonyan2014very,alshareef2026end}.

\textbf{Vision transformers (ViTs)} have significantly advanced deep learning in computer vision. Compared to Convolutional Neural Networks (CNNs), Vision Transformer (ViTs) utilize self-attention to extract local and global characteristics from image input, which are then fed directly into a fully networked multilayer perceptron head via residual connections \cite{abou2023white}. Google's Vision Transformer is a transformer encoder-based architecture that converts an image into a series of fixed-size picture patches that are linearly embedded and processed utilizing self-attention techniques. In this study, we use a ViT model pretrained on ImageNet-21K and subsequently fine-tuned on ImageNet-1K~\cite{dosovitskiy2020image}.

\textbf{Data-efficient Image Transformer (DeiT)} is a ViT architecture designed for efficient training on ImageNet-1K without large-scale external pretraining. It introduces a distillation token that enables the transformer to learn from a CNN teacher while preserving its attention-based architecture, achieving competitive image classification performance with lower training requirements~\cite{touvron2021training}.

\textbf{Swin Transformer} is a hierarchical vision transformer that applies self-attention within local windows and uses shifted windows to exchange information across regions. This design supports multi-scale feature extraction with computational complexity that scales linearly with image size~\cite{liu2021swin}.

\textbf{Implementation environment.}
The experiments were conducted on computing nodes equipped with either an NVIDIA H200 GPU with 141~GB of memory or an NVIDIA A100 GPU with 40~GB of memory. APQF was implemented in Python~3.11.6 using PyTorch~2.6.0 with CUDA~12.4.

\subsection{Results on ImageNet}
We evaluate APQF on ImageNet using two metrics. The first is Top-1 accuracy on the validation set, which we report as the base accuracy of the original model, the final accuracy after compression, and their difference $\Delta$, so the accuracy cost of compression is directly visible. The second is the relative bit-operations (Rel. BOPs), the ratio of the compressed model's bit-operations to those of the original FP32 model. Since BOPs weight each layer's multiply--accumulate operations by its weight and activation bit-widths, Rel. BOPs captures both the pruning and the quantization gains in a single number, where lower is better.

The results here use a deliberately constrained setting. Each model is trained on only a 200K-image subset of ImageNet, using 200 images per class, and evaluated on the full 50{,}000-image validation set. The quantization agent uses its conservative prompt, and the optional post-quantization knowledge-distillation stage is disabled and left as a source of further improvement.

\begin{table}[!htbp] \centering
\isucaption{APQF compression results on ImageNet (conservative mixed-precision QAT on pruned models; 200K training / 50K validation samples).}
\label{tab:imagenet_results}
\begin{tabular}{l|l|cccc} \hline
\textbf{Dataset} & \textbf{Model} & \textbf{Base Top-1 (\%)} & \textbf{Final Top-1 (\%)} & \textbf{$\Delta$ Top-1} & \textbf{Rel. BOPs (\%)} \\ \hline
\multirow{4}{*}{ImageNet}
 & DeiT-Tiny  & 66.52 & 67.90 & $+1.38$ & 5.63 \\
 & ResNet-50  & 80.13 & 72.73 & $-7.40$ & 7.74 \\
 & Swin-Tiny  & 80.61 & 74.16 & $-6.45$ & 6.17 \\
 & ViT-Small  & 79.87 & 76.50 & $-3.37$ & 5.93 \\ \hline
\end{tabular}
\end{table}

Table~\ref{tab:imagenet_results} reports four architectures covering both CNN and transformer families. Across all of them APQF cuts the compute cost to about 5.6--7.7\% of the original, roughly a 13--18$\times$ reduction in bit-operations, while keeping accuracy close to the baseline. DeiT-Tiny is only lightly pruned and even improves ($+1.38$), whereas the more heavily compressed ResNet-50 and Swin-Tiny show larger but still moderate drops. Given the small training budget, the conservative quantization, and the absence of post-QAT recovery, these results show that APQF reaches large compute reductions at a modest accuracy cost.

\begin{figure}[!htbp] \centering
\includegraphics[width=0.7\textwidth]{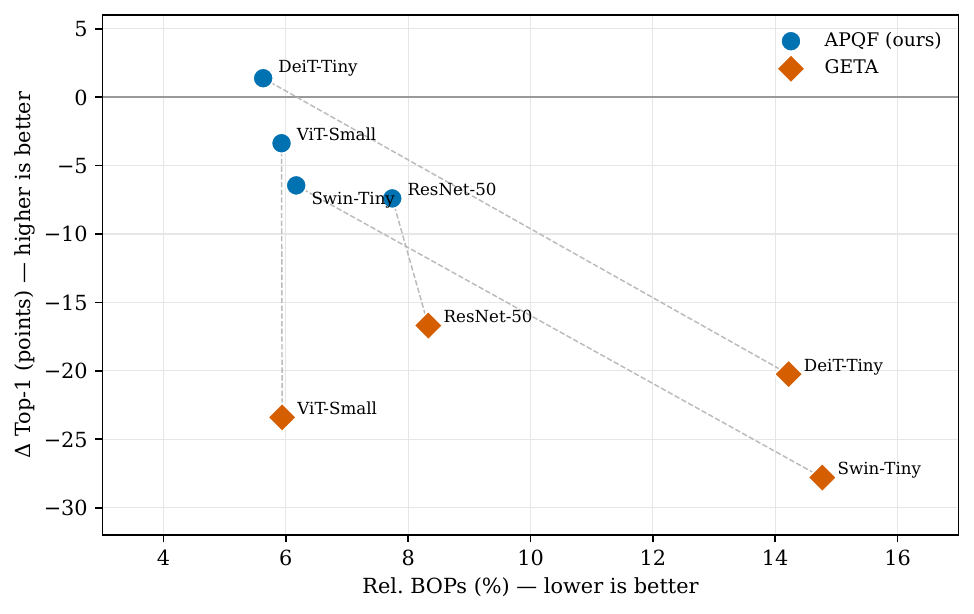}
\caption{Accuracy--compute trade-off on ImageNet under a 200K training budget. Each model has an APQF point (circle) and a GETA point (diamond) joined by a dashed line. APQF attains near-zero accuracy loss at low relative BOPs, whereas GETA degrades substantially, reflecting its higher data requirement}
\label{fig:geta_tradeoff}
\end{figure}

As shown in Figure~\ref{fig:geta_tradeoff}, under a constrained training budget of 200K images, GETA's~\cite{qu2025automatic} single-run joint pruning--quantization optimization fails to recover accuracy, dropping to 51--59\% Top-1, whereas APQF retains 68--77\%. For a fair comparison, we run GETA ourselves using the same configuration provided in their released code. The gap reflects a fundamental difference in how the two methods reach their compressed models. Although both start from the same pretrained weights, GETA re-optimizes the entire network into a sparse and quantized subnet through its joint optimizer, which needs large-scale training data to converge. In contrast, APQF prunes the pretrained model directly from measured sensitivity and recovers accuracy through knowledge distillation and parameter-efficient fine-tuning, making it substantially more data-efficient

Table~\ref{tab:geta_fulldata} compares the two methods when both have access to the full ImageNet training set. GETA is trained on all of the data for its full training schedule, following its released configuration. APQF instead reuses the model produced by its limited-data pipeline and simply activates the optional post-quantization knowledge-distillation stage for only around five epochs on the full training set. Even with this short additional recovery, APQF matches GETA's accuracy to within a few points while achieving roughly a $3\times$ lower relative BOPs, because APQF combines structured pruning with aggressive mixed-precision quantization whereas GETA relies primarily on structured pruning.

\begin{table}[!htbp] \centering
\isucaption{APQF vs.\ GETA on ImageNet with full-data training, where APQF additionally uses the optional post-quantization KD stage ($\sim$5 epochs).}
\label{tab:geta_fulldata}
\begin{tabular}{l|l|ccc} \hline
\textbf{Model} & \textbf{Method} & \textbf{Base $\rightarrow$ Final (\%)} & \textbf{$\Delta$ Top-1} & \textbf{Rel. BOPs (\%)} \\ \hline
\multirow{2}{*}{ViT-Small} & GETA & $81.43 \rightarrow 80.12$ & $-1.31$ & 19.37 \\
                           & APQF & $79.87 \rightarrow 78.26$ & $-1.61$ & \textbf{5.73} \\ \hline
\multirow{2}{*}{DeiT-Tiny} & GETA & $72.01 \rightarrow 72.88$ & $+0.87$ & 16.95 \\
                           & APQF & $66.52 \rightarrow 69.51$ & $\mathbf{+2.99}$ & \textbf{6.11} \\ \hline
\multirow{2}{*}{Swin-Tiny} & GETA & $80.92 \rightarrow 80.09$ & $-0.83$ & 21.84 \\
                           & APQF & $80.61 \rightarrow 77.59$ & $-3.02$ & \textbf{7.23} \\ \hline
\end{tabular}
\end{table}

We compare against GETA using the authors' reproducible baseline configuration for each architecture. For the released GETA implementation, weight-only quantization is the stable and general path for ResNet and transformer backbones. We attempted to enable activation quantization in GETA, but doing so was not a simple flag change, and it exposed pruning-dependency and subnet-construction failures for ResNet and ViT-style models. Enabling joint weight and activation quantization in GETA would therefore require nontrivial modifications to the baseline implementation, which would no longer be the authors' original method.

In contrast, our method supports joint pruning with both weight and activation quantization across these architectures. Since activation quantization usually makes optimization harder and can reduce accuracy, achieving better accuracy and Rel. BOPs while quantizing activations is a stronger result than comparing only weight quantization.

\subsection{Results on CIFAR-10}
Table~\ref{tab:geta_cifar} compares APQF and GETA on CIFAR-10, where both methods start from identical fine-tuned checkpoints so the baselines are matched.The small differences in base accuracy arise because each method measures accuracy with its own evaluation pipeline, using different preprocessing and validation sampling. Since $\Delta$ is computed relative to each method's own measured baseline, these differences do not affect the fairness of the comparison, as each method's accuracy change is assessed against its own starting point. Across the five architectures, APQF compresses more while retaining higher accuracy in most cases. It reaches lower relative BOPs on four of the five models and improves over its own baseline on three of them, whereas GETA loses several points of accuracy on ResNet-50, Swin-Tiny, and ViT-Small. GETA is ahead only on ResNet-20 in accuracy and on ViT-Small in relative BOPs.

\begin{table}[!htbp] \centering
\isucaption{APQF vs.\ GETA on CIFAR-10, using identical finetuned checkpoints (baseline-matched). $\Delta$ is the accuracy change relative to each method's own baseline; bold marks the better value.}
\label{tab:geta_cifar}
\begin{tabular}{l|l|ccc} \hline
\textbf{Model} & \textbf{Method} & \textbf{Base $\rightarrow$ Final (\%)} & \textbf{$\Delta$ Top-1} & \textbf{Rel. BOPs (\%)} \\ \hline
\multirow{2}{*}{DeiT-Tiny} & GETA & $91.75 \rightarrow 91.05$ & $-0.70$ & 11.90 \\
                           & APQF & $91.90 \rightarrow 95.01$ & $\mathbf{+3.11}$ & \textbf{8.49} \\ \hline
\multirow{2}{*}{ResNet-50} & GETA & $96.55 \rightarrow 93.09$ & $-3.46$ & 7.68 \\
                           & APQF & $95.83 \rightarrow 96.62$ & $\mathbf{+0.79}$ & \textbf{7.65} \\ \hline
\multirow{2}{*}{Swin-Tiny} & GETA & $97.20 \rightarrow 91.46$ & $-5.74$ & 15.12 \\
                           & APQF & $96.78 \rightarrow 97.92$ & $\mathbf{+1.14}$ & \textbf{5.11} \\ \hline
\multirow{2}{*}{ViT-Small} & GETA & $98.51 \rightarrow 92.16$ & $-6.35$ & \textbf{3.59} \\
                           & APQF & $98.46 \rightarrow 98.27$ & $\mathbf{-0.19}$ & 4.27 \\ \hline
\multirow{2}{*}{ResNet-20} & GETA & $91.70 \rightarrow 91.42$ & $\mathbf{-0.28}$ & 4.50 \\
                           & APQF & $92.12 \rightarrow 91.55$ & $-0.57$ & \textbf{2.90} \\ \hline
\end{tabular}
\end{table}

\subsubsection{VGG7 on CIFAR-10}
Table~\ref{tab:vgg7} compares APQF with quantization-only and combined pruning--quantization methods on VGG7/CIFAR-10. The results for the comparison methods are taken from the corresponding publications. APQF is evaluated using our FP32 VGG7 reproduction, which achieves \(92.95\%\) accuracy. Therefore, its \(\Delta\)Acc and relative BOPs are calculated with respect to this baseline, whereas the corresponding values for prior methods use their reported \(93.05\%\) baseline.
Among the methods combining pruning and quantization, APQF provides a favorable accuracy--efficiency trade-off. At the same reported relative BOP budget of \(0.41\%\), APQF achieves \(93.15\%\) accuracy compared with \(92.57\%\) for GETA, an improvement of \(0.58\) percentage points. At a lower budget of \(0.36\%\), APQF still exceeds GETA by \(0.11\) points while requiring fewer bit-operations.
The highest-accuracy Bayesian Bits configuration achieves \(93.23\%\) accuracy at \(0.51\%\) relative BOPs. APQF achieves an accuracy within \(0.08\) percentage points of this result while using approximately \(20\%\) fewer bit-operations (\(0.41\%\) versus \(0.51\%\)). The lower-cost Bayesian Bits configuration reduces relative BOPs to \(0.29\%\), but its accuracy decreases by \(1.27\) points relative to its higher-accuracy configuration. APQF is also the only method operating below \(0.5\%\) relative BOPs that improves upon its own FP32 baseline, achieving \(\Delta\text{Acc}=+0.20\).
The conventional quantization-only baselines, including TWN, LR-Net, RQ, and WAGE, require approximately \(1.56\%\)--\(6.24\%\) relative BOPs, corresponding to approximately \(4\)--\(17\) times the cost of the APQF configurations. DQ reaches a comparable BOP range of \(0.48\%\)--\(0.54\%\), but with substantially lower accuracy. Overall, these results demonstrate the benefit of combining structured pruning with mixed-precision quantization.

\begin{table}[!htbp]
\centering
\small
\isucaption{VGG7 on CIFAR-10: comparison of quantization and joint pruning--quantization methods. A \checkmark\ in ``Prune'' denotes structured pruning, while $\times$ indicates that pruning is not applied. ``Quant.'' lists the quantization scheme (W = weights, A = activations). $\Delta$Acc is the change from each method's FP32 baseline; Rel.\ BOPs is relative to that baseline (lower is better).}
\label{tab:vgg7}

\setlength{\tabcolsep}{4pt}
\begin{tabular}{l|c l ccc}
\hline
\textbf{Method} & \textbf{Prune} & \textbf{Quant.} &
\textbf{Acc. (\%)} & \textbf{$\Delta$Acc} &
\textbf{Rel. BOPs (\%)} \\
\hline

Baseline
& -- & -- & 93.05 & -- & 100.00 \\
\hline

TWN~\cite{li2016ternary}
& $\times$ & \checkmark\,(ternary W, 2/32)
& 92.56 & $-0.49$ & 6.24 \\

LR-Net~\cite{shayer2017learning}
& $\times$ & \checkmark\,(1-bit W)
& 93.18 & $+0.13$ & 3.11 \\

RQ (8/8)~\cite{louizos2018relaxed}
& $\times$ & \checkmark\,(W+A, 8/8)
& 93.30 & $+0.25$ & 6.24 \\

RQ (4/4)~\cite{louizos2018relaxed}
& $\times$ & \checkmark\,(W+A, 4/4)
& 92.04 & $-1.01$ & 1.56 \\

WAGE~\cite{wu2018training}
& $\times$ & \checkmark\,(W+A, 2/8)
& 93.22 & $+0.17$ & 1.56 \\

DQ~\cite{uhlich2019mixed}
& $\times$ & \checkmark\,(mixed W+A)
& 91.59 & $-1.46$ & 0.48 \\

DQ-restrict~\cite{uhlich2019mixed}
& $\times$ & \checkmark\,(mixed W+A)
& 91.59 & $-1.46$ & 0.54 \\

DJPQ~\cite{wang2020differentiable}
& \checkmark & \checkmark\,(mixed W+A)
& 91.54 & $-1.51$ & 0.48 \\

DJPQ-restrict~\cite{wang2020differentiable}
& \checkmark & \checkmark\,(mixed W+A)
& 91.43 & $-1.62$ & 0.46 \\

BB ($\mu{=}0.01$)~\cite{van2020bayesian}
& \checkmark & \checkmark\,(mixed W+A)
& 93.23 & $+0.18$ & 0.51 \\

BB ($\mu{=}0.1$)~\cite{van2020bayesian}
& \checkmark & \checkmark\,(mixed W+A)
& 91.96 & $-1.09$ & \textbf{0.29} \\

GETA~\cite{qu2025automatic}
& \checkmark & \checkmark\,(mixed W+A)
& 92.57 & $-0.48$ & 0.41 \\

APQF (ours)$^{\dagger}$
& \checkmark & \checkmark\,(mixed W+A)
& 92.68 & $-0.27$ & 0.36 \\

APQF (ours)$^{\dagger}$
& \checkmark & \checkmark\,(mixed W+A)
& \textbf{93.15} & $\mathbf{+0.20}$ & 0.41 \\
\hline
\end{tabular}

\vspace{2pt}
{\footnotesize
$^{\dagger}$APQF is compressed from our VGG7 reproduction (92.95\% FP32);
$\Delta$Acc and Rel.\ BOPs are relative to that baseline.
}
\end{table}

\subsubsection{Ablation Study}
We conduct an ablation study on CIFAR-10 to isolate the contribution of the main components of APQF, shown in Figure~\ref{fig:ablation_cifar}. The full method is compared against two reduced variants at roughly matched relative BOPs, so accuracy is compared at equal compute. The first variant, Uniform, removes the LLM's adaptive per-layer decisions and instead applies standard dependency-graph (DepGraph) pruning~\cite{fang2023depgraph} followed by Brevitas quantization~\cite{brevitas}. It uses a single pruning ratio for every layer and one fixed bit-width across the whole model, set so that its relative BOPs matches the full method. The second variant, w/o profiling, still uses the LLM to produce the pruning and quantization strategy, but the LLM no longer receives the measured profiling data. It therefore makes its decisions from its own prior knowledge of the architecture rather than from measured sensitivity. Across all four models, Uniform causes the largest accuracy drop, confirming that the LLM's per-layer pruning ratios and mixed precision are the main drivers of accuracy at a given compute budget, while removing the profiling data causes a smaller but consistent drop, showing that grounding the LLM in measured sensitivity further improves its decisions. Together, these results show that the different parts of APQF each contribute to its performance.
\definecolor{apqfblue}{RGB}{42,120,214}
\definecolor{unifred}{RGB}{227,73,72}
\definecolor{noprofor}{RGB}{237,161,0}
\begin{figure}[!htbp] \centering
\includegraphics[width=0.8\textwidth]{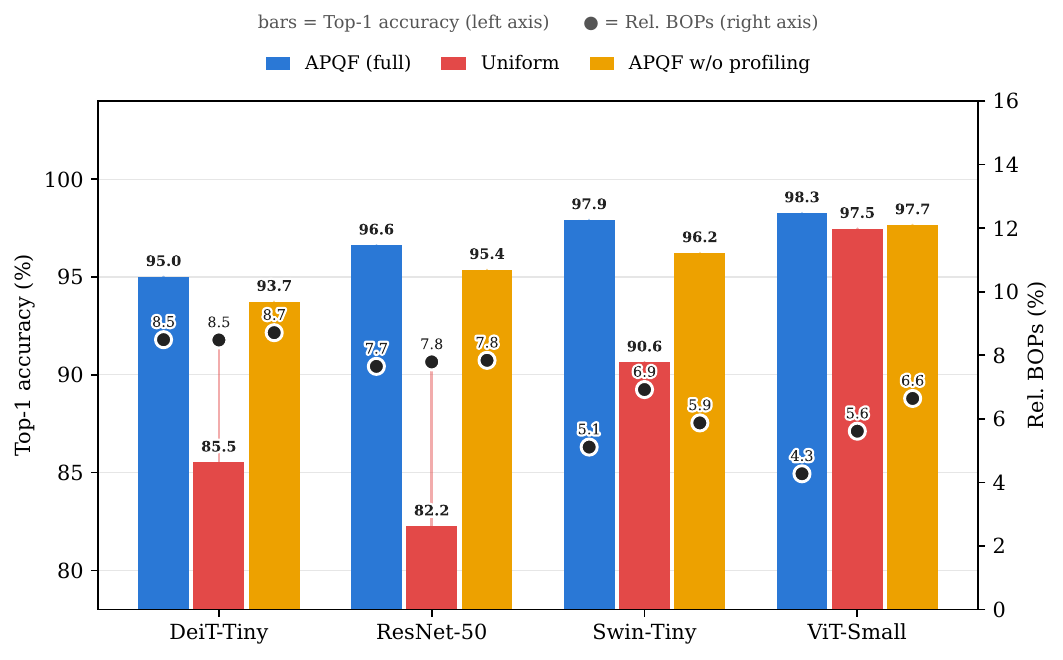}
\caption{CIFAR-10 ablation. Bars are Top-1 accuracy (left axis); dots are Rel.\ BOPs (right axis). \textcolor{apqfblue}{\textbf{APQF (full)}} uses per-layer pruning ratios and mixed precision, \textcolor{unifred} {\textbf{Uniform}} uses one ratio and one bit-width for all layers, and \textcolor{noprofor} {\textbf{APQF w/o profiling}} removes the profiling agent. At matched compute, Uniform loses the most accuracy while removing profiling costs less. (Left axis from 78\%.)}
\label{fig:ablation_cifar}
\end{figure}

\subsubsection{Robustness to the LLM Planner}
\begin{figure}[!htbp] \centering
\includegraphics[width=0.7\textwidth]{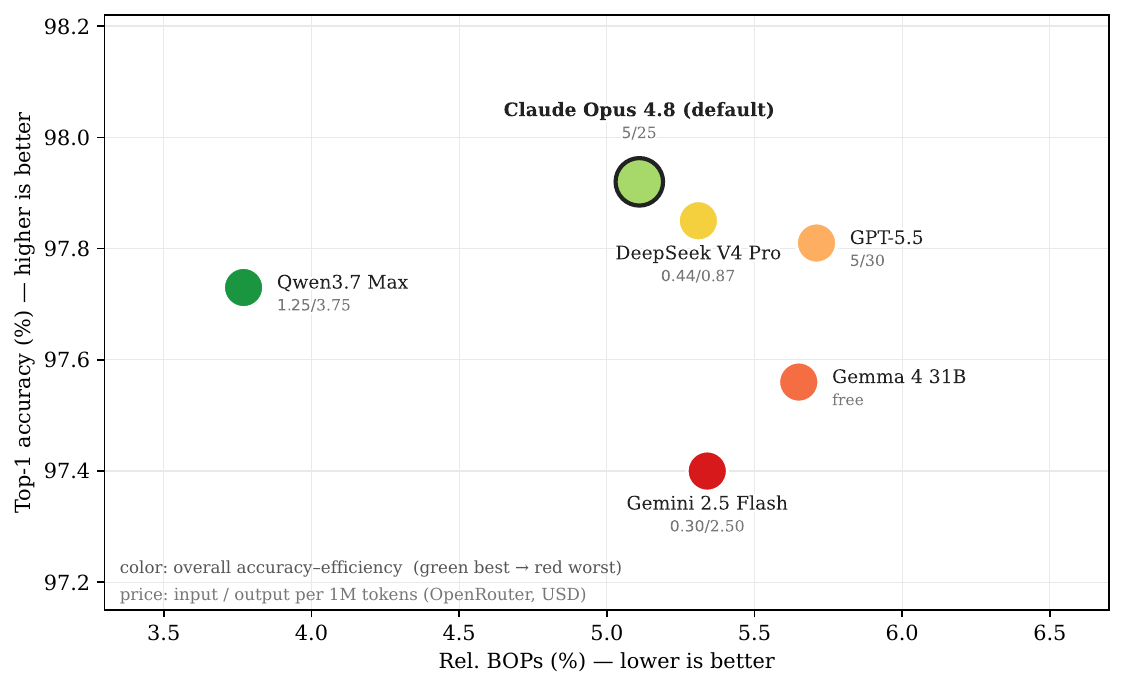}
\caption{Effect of the LLM planner (Swin-Tiny, CIFAR-10). Each point is one LLM, colored by accuracy--efficiency ranking (green best, red worst) and labeled with its OpenRouter price per 1M tokens. Accuracy stays stable across a wide cost range}
\label{fig:llm_comparison}
\end{figure}

To test how much APQF depends on the specific model used as its LLM planner, we run the full pipeline on Swin-Tiny with six different LLMs while keeping everything else fixed. Figure~\ref{fig:llm_comparison} plots the resulting accuracy against relative BOPs, with each point colored by an overall accuracy-efficiency ranking and labeled with its OpenRouter price. Accuracy stays within a narrow band of 97.4 to 97.9 percent across all six models, while the relative BOPs and the cost per token vary substantially. This shows that APQF is robust to the choice of LLM and does not rely on a single premium model. It performs well even with inexpensive or free planners such as DeepSeek and Gemma, which reach accuracy comparable to the much more expensive Claude Opus 4.8 and GPT-5.5. Claude Opus 4.8, used by default in our main experiments, attains the highest accuracy, while Qwen3.7 Max and DeepSeek V4 Pro reach comparable accuracy at a fraction of the cost.

\section{Summary and Discussion}
Vision models are now used across a wide range of domains, but their size makes them expensive to adapt and deploy, and their inference latency and memory footprint often exceed what resource-constrained hardware can provide. Compressing them is therefore necessary, yet deciding how to compress a
particular model is still largely a manual process built on expert judgment and repeated trial and error, and settings that work well for one architecture seldom transfer to another.

We introduced APQF, an agentic, profiling-guided framework that combines
structured pruning, mixed-precision quantization, and accuracy recovery in a single automated pipeline. Rather than applying fixed heuristics or one uniform rule to every layer, APQF harnesses the reasoning ability of large language models to plan compression from measured profiling evidence, so that the per-layer pruning ratios, the per-layer bit-widths, and the recovery strategy are decided
automatically instead of being hand-tuned. To our knowledge, this is the first framework to combine LLM-guided, profiling-grounded decisions with a fully training-aware pipeline that performs sequential structured pruning together with mixed-precision quantization-aware training and recovery.

Five cooperating agents carry this out. The ProfilingAgent produces an
architecture brief and a pruning sensitivity map that ground every later decision in measurement. The PrunerAgent performs multi-stage structured pruning toward the user's parameter-reduction target, calling the FineTuningAgent to recover accuracy between stages, and the QuantAgent then assigns per-layer bit-widths and trains the quantized model with knowledge distillation from a full-precision teacher. The EvaluationAgent measures every checkpoint on the same fixed
validation set, and every strategy an LLM proposes is validated against the actual model before it is executed.

We demonstrated that basing the compression decisions on measured profiling evidence yields a better accuracy-efficiency trade-off than uniform compression, and that the same pipeline applies to different architectures without modification, since only the profiling data changes from one model to the next. Compared against established non-agentic compression frameworks on models ranging from convolutional networks to vision transformers, APQF matched or exceeded their
accuracy in most settings while producing models with lower computational cost. Because APQF compresses a pretrained model by recovering it rather than re-optimizing the whole network, it retains considerably more accuracy than joint optimization methods when the training budget is limited. The planning itself is
performed through a standard API, so the framework is not tied to any single language model. Our experiments show that it works with freely available open-weight planners as well as with the most expensive commercial ones, and the choice is left to the user, so APQF can also benefit directly from stronger planners as they become available. The same approach could be extended to further compression techniques and to tasks beyond image classification.

\section{Limitations and Future Work}
APQF currently targets image classification models,
and although it covers both convolutional networks and vision transformers, the idea itself is not specific to classification. We intend to apply the same approach to other foundation models, such as the Segment Anything Model for segmentation, and to models outside vision. We also believe it can be extended to compress large language models and vision-language models, since nothing in the profiling and planning mechanism depends on the task, and we see clear potential for it to expand to further model families and datasets.

Using language models as planners carries a limitation inherent to them, since the same request can yield a different plan on a different run. We limit this with carefully designed prompts and by controlling the sampling temperature, but the planner can still occasionally fail to produce a usable strategy. Such failures were rare with strong models and more frequent with cheap or free ones, and also in the ablation where the profiling evidence was withheld, which suggests the planner relies on that evidence to stay within a valid design space. Better prompting and stronger planners would reduce this further. Strong commercial models are expensive, but as our experiments show, APQF is not tied to any single one and can take advantage of cheaper models as well as of stronger ones that appear in the future.

Compression is most valuable on large models, but pruning, fine-tuning, and quantization-aware training on those models are themselves computationally expensive. This is especially true in vision, where the experiments run on large datasets such as ImageNet, and where a single configuration involves repeated rounds of pruning, recovery, and quantization-aware training. Carrying out this research therefore required multiple GPUs, and the available compute limited how
many configurations and repeated runs we could explore.

Some human involvement also remains. The user still selects which language model to use for which agent, a few of its sampling settings, and some distillation hyperparameters, and to keep the comparison with other methods fair the user also specifies how much to prune and how aggressively to quantize. These choices are made through the configuration file, and in practice they also help the planner produce valid pruning, quantization, and recovery strategies, since leaving them entirely open raises the failure rate.

\section*{Acknowledgment}
This work used GPU resources on the NCSA Delta GPU system, provided through ACCESS allocations AGR250012, CIS250538, and CIS240379, which made it possible to carry out the
experiments presented here. We thank Dr.\ Christopher Quinn for his guidance and support, and Aishwarya Sarkar, PhD candidate, for her help during the early stages of this work.

\bibliographystyle{unsrt}  
\bibliography{references}

@article{zhu2025comprehensive,
  title={A comprehensive review of network pruning based on pruning granularity and pruning time perspectives},
  author={Zhu, Kehan and Hu, Fuyi and Ding, Yuanbing and Zhou, Wei and Wang, Ruxin},
  journal={Neurocomputing},
  volume={626},
  pages={129382},
  year={2025},
  publisher={Elsevier}
}

@article{ngo2025edge,
  title={Edge intelligence: A review of deep neural network inference in resource-limited environments},
  author={Ngo, Dat and Park, Hyun-Cheol and Kang, Bongsoon},
  journal={Electronics},
  volume={14},
  number={12},
  pages={2495},
  year={2025},
  publisher={MDPI}
}

@article{rafat2023mitigating,
  title={Mitigating carbon footprint for knowledge distillation based deep learning model compression},
  author={Rafat, Kazi and Islam, Sadia and Mahfug, Abdullah Al and Hossain, Md Ismail and Rahman, Fuad and Momen, Sifat and Rahman, Shafin and Mohammed, Nabeel},
  journal={Plos one},
  volume={18},
  number={5},
  pages={e0285668},
  year={2023},
  publisher={Public Library of Science San Francisco, CA USA}
}

@article{malihi2023efficient,
  title={Efficient and controllable model compression through sequential knowledge distillation and pruning},
  author={Malihi, Leila and Heidemann, Gunther},
  journal={Big Data and Cognitive Computing},
  volume={7},
  number={3},
  pages={154},
  year={2023},
  publisher={MDPI}
}

@article{argerich2024measuring,
  title={Measuring and improving the energy efficiency of large language models inference},
  author={Argerich, Mauricio Fadel and Pati{\~n}o-Mart{\'\i}nez, Marta},
  journal={IEEE Access},
  volume={12},
  pages={80194--80207},
  year={2024},
  publisher={IEEE}
}

@article{xu2026parameter,
  title={Parameter-efficient fine-tuning methods for pretrained language models: A critical review and assessment},
  author={Xu, Lingling and Xie, Haoran and Qin, S Joe and Tao, Xiaohui and Wang, Fu Lee},
  journal={IEEE Transactions on Pattern Analysis and Machine Intelligence},
  year={2026},
  publisher={IEEE}
}

@article{li2023model,
  title={Model compression for deep neural networks: A survey},
  author={Li, Zhuo and Li, Hengyi and Meng, Lin},
  journal={Computers},
  volume={12},
  number={3},
  pages={60},
  year={2023},
  publisher={MDPI}
}

@article{han2015learning,
  title={Learning both weights and connections for efficient neural network},
  author={Han, Song and Pool, Jeff and Tran, John and Dally, William},
  journal={Advances in neural information processing systems},
  volume={28},
  year={2015}
}

@article{filters2016pruning,
  title={Pruning filters for efficient convnets},
  author={Filters’Importance, Determine},
  journal={arXiv preprint arXiv:1608.08710},
  volume={3},
  year={2016}
}

@article{cheng2024survey,
  title={A survey on deep neural network pruning: Taxonomy, comparison, analysis, and recommendations},
  author={Cheng, Hongrong and Zhang, Miao and Shi, Javen Qinfeng},
  journal={IEEE Transactions on Pattern Analysis and Machine Intelligence},
  volume={46},
  number={12},
  pages={10558--10578},
  year={2024},
  publisher={IEEE}
}

@inproceedings{frantar2023sparsegpt,
  title={Sparsegpt: Massive language models can be accurately pruned in one-shot},
  author={Frantar, Elias and Alistarh, Dan},
  booktitle={International conference on machine learning},
  pages={10323--10337},
  year={2023},
  organization={PMLR}
}

@article{he2023structured,
  title={Structured pruning for deep convolutional neural networks: A survey},
  author={He, Yang and Xiao, Lingao},
  journal={IEEE transactions on pattern analysis and machine intelligence},
  volume={46},
  number={5},
  pages={2900--2919},
  year={2023},
  publisher={IEEE}
}

@article{zhou2017incremental,
  title={Incremental network quantization: Towards lossless cnns with low-precision weights},
  author={Zhou, Aojun and Yao, Anbang and Guo, Yiwen and Xu, Lin and Chen, Yurong},
  journal={arXiv preprint arXiv:1702.03044},
  year={2017}
}

@inproceedings{zhao2023post,
  title={Post-training quantization or quantization-aware training? That is the question},
  author={Zhao, Xiaotian and Xu, Ruge and Guo, Xinfei},
  booktitle={2023 China Semiconductor Technology International Conference (CSTIC)},
  pages={1--3},
  year={2023},
  organization={IEEE}
}

@inproceedings{liu2023pd,
  title={Pd-quant: Post-training quantization based on prediction difference metric},
  author={Liu, Jiawei and Niu, Lin and Yuan, Zhihang and Yang, Dawei and Wang, Xinggang and Liu, Wenyu},
  booktitle={Proceedings of the IEEE/CVF Conference on Computer Vision and Pattern Recognition},
  pages={24427--24437},
  year={2023}
}

@inproceedings{chen2025efficientqat,
  title={Efficientqat: Efficient quantization-aware training for large language models},
  author={Chen, Mengzhao and Shao, Wenqi and Xu, Peng and Wang, Jiahao and Gao, Peng and Zhang, Kaipeng and Luo, Ping},
  booktitle={Proceedings of the 63rd Annual Meeting of the Association for Computational Linguistics (Volume 1: Long Papers)},
  pages={10081--10100},
  year={2025}
}

@phdthesis{xu2026deep,
  title={Deep model compression via knowledge distillation for time-series data analytics: from in-domain to cross-domain scenarios},
  author={Xu, Qing},
  year={2026},
  school={Nanyang Technological University}
}

@article{yang2025feature,
  title={Feature alignment and representation transfer in knowledge distillation for large language models},
  author={Yang, Junjie and Song, Junhao and Han, Xudong and Bi, Ziqian and Wang, Tianyang and Liang, Chia Xin and Song, Xinyuan and Zhang, Yichao and Niu, Qian and Peng, Benji and others},
  journal={arXiv preprint arXiv:2504.13825},
  year={2025}
}

@article{moslemi2024survey,
  title={A survey on knowledge distillation: Recent advancements},
  author={Moslemi, Amir and Briskina, Anna and Dang, Zubeka and Li, Jason},
  journal={Machine Learning with Applications},
  volume={18},
  pages={100605},
  year={2024},
  publisher={Elsevier}
}

@article{wang2025parameter,
  title={Parameter-efficient fine-tuning in large language models: a survey of methodologies},
  author={Wang, Luping and Chen, Sheng and Jiang, Linnan and Pan, Shu and Cai, Runze and Yang, Sen and Yang, Fei},
  journal={Artificial Intelligence Review},
  volume={58},
  number={8},
  pages={227},
  year={2025},
  publisher={Springer}
}

@article{zhang2025parameter,
  title={Parameter-efficient fine-tuning for foundation models},
  author={Zhang, Dan and Feng, Tao and Xue, Lilong and Wang, Yuandong and Dong, Yuxiao and Tang, Jie},
  journal={arXiv preprint arXiv:2501.13787},
  year={2025}
}

@article{yuan2025efficientllm,
  title={EfficientLLM: Efficiency in Large Language Models},
  author={Yuan, Zhengqing and Sun, Weixiang and Liu, Yixin and Zhou, Huichi and Zhou, Rong and Li, Yiyang and Zhang, Zheyuan and Song, Wei and Huang, Yue and Jia, Haolong and others},
  journal={arXiv preprint arXiv:2505.13840},
  year={2025}
}

@article{hu2022lora,
  title={Lora: Low-rank adaptation of large language models.},
  author={Hu, Edward J and Shen, Yelong and Wallis, Phillip and Allen-Zhu, Zeyuan and Li, Yuanzhi and Wang, Shean and Wang, Liang and Chen, Weizhu and others},
  journal={Iclr},
  volume={1},
  number={2},
  pages={3},
  year={2022}
}

@inproceedings{liu2024dora,
  title={Dora: Weight-decomposed low-rank adaptation},
  author={Liu, Shih-Yang and Wang, Chien-Yi and Yin, Hongxu and Molchanov, Pavlo and Wang, Yu-Chiang Frank and Cheng, Kwang-Ting and Chen, Min-Hung},
  booktitle={Forty-first International Conference on Machine Learning},
  year={2024}
}

@article{meng2024pissa,
  title={Pissa: Principal singular values and singular vectors adaptation of large language models},
  author={Meng, Fanxu and Wang, Zhaohui and Zhang, Muhan},
  journal={Advances in Neural Information Processing Systems},
  volume={37},
  pages={121038--121072},
  year={2024}
}

@article{kalajdzievski2023rank,
  title={A rank stabilization scaling factor for fine-tuning with lora},
  author={Kalajdzievski, Damjan},
  journal={arXiv preprint arXiv:2312.03732},
  year={2023}
}

@article{hayou2024lora+,
  title={Lora+: Efficient low rank adaptation of large models},
  author={Hayou, Soufiane and Ghosh, Nikhil and Yu, Bin},
  journal={arXiv preprint arXiv:2402.12354},
  year={2024}
}

@article{wu2025agentic,
  title={Agentic reasoning: Reasoning llms with tools for the deep research},
  author={Wu, Junde and Zhu, Jiayuan and Liu, Yuyuan},
  journal={arXiv preprint arXiv:2502.04644},
  volume={9},
  year={2025}
}

@article{zhao2025llm,
  title={Llm-based agentic reasoning frameworks: A survey from methods to scenarios},
  author={Zhao, Bingxi and Foo, Lin Geng and Hu, Ping and Theobalt, Christian and Rahmani, Hossein and Liu, Jun},
  journal={arXiv preprint arXiv:2508.17692},
  year={2025}
}

@inproceedings{prajwal2023comparative,
  title={A comparative study of resnet-pretrained models for computer vision},
  author={Prajwal, Thode Sai and AK, Ilavarasi},
  booktitle={Proceedings of the 2023 Fifteenth International Conference on Contemporary Computing},
  pages={419--425},
  year={2023}
}

@inproceedings{he2016deep,
  title={Deep residual learning for image recognition},
  author={He, Kaiming and Zhang, Xiangyu and Ren, Shaoqing and Sun, Jian},
  booktitle={Proceedings of the IEEE conference on computer vision and pattern recognition},
  pages={770--778},
  year={2016}
}

@article{abou2023white,
  title={White blood cell classification: Convolutional Neural Network (CNN) and Vision Transformer (ViT) under medical microscope},
  author={Abou Ali, Mohamad and Dornaika, Fadi and Arganda-Carreras, Ignacio},
  journal={Algorithms},
  volume={16},
  number={11},
  pages={525},
  year={2023},
  publisher={MDPI}
}

@article{dosovitskiy2020image,
  title={An image is worth 16x16 words: Transformers for image recognition at scale},
  author={Dosovitskiy, Alexey and Beyer, Lucas and Kolesnikov, Alexander and Weissenborn, Dirk and Zhai, Xiaohua and Unterthiner, Thomas and Dehghani, Mostafa and Minderer, Matthias and Heigold, Georg and Gelly, Sylvain and others},
  journal={arXiv preprint arXiv:2010.11929},
  year={2020}
}

@inproceedings{liu2021swin,
  title={Swin transformer: Hierarchical vision transformer using shifted windows},
  author={Liu, Ze and Lin, Yutong and Cao, Yue and Hu, Han and Wei, Yixuan and Zhang, Zheng and Lin, Stephen and Guo, Baining},
  booktitle={Proceedings of the IEEE/CVF international conference on computer vision},
  pages={10012--10022},
  year={2021}
}

@inproceedings{touvron2021training,
  title={Training data-efficient image transformers \& distillation through attention},
  author={Touvron, Hugo and Cord, Matthieu and Douze, Matthijs and Massa, Francisco and Sablayrolles, Alexandre and J{\'e}gou, Herv{\'e}},
  booktitle={International conference on machine learning},
  pages={10347--10357},
  year={2021},
  organization={PMLR}
}

@article{simonyan2014very,
  title={Very deep convolutional networks for large-scale image recognition},
  author={Simonyan, Karen and Zisserman, Andrew},
  journal={arXiv preprint arXiv:1409.1556},
  year={2014}
}

@article{alshareef2026end,
  title={End-to-end discrete cosine transform integration in spectral convolutional neural networks for resource-efficient deep learning},
  author={Alshareef, Ibrahim Yousef and Ab Rahman, Ab Al-Hadi and Khan, Nuzhat and Alqaraghuli, Hasan},
  journal={Applied Soft Computing},
  pages={114599},
  year={2026},
  publisher={Elsevier}
}

@inproceedings{deng2009imagenet,
  title={Imagenet: A large-scale hierarchical image database},
  author={Deng, Jia and Dong, Wei and Socher, Richard and Li, Li-Jia and Li, Kai and Fei-Fei, Li},
  booktitle={2009 IEEE conference on computer vision and pattern recognition},
  pages={248--255},
  year={2009},
  organization={Ieee}
}

@article{russakovsky2015imagenet,
  title={Imagenet large scale visual recognition challenge},
  author={Russakovsky, Olga and Deng, Jia and Su, Hao and Krause, Jonathan and Satheesh, Sanjeev and Ma, Sean and Huang, Zhiheng and Karpathy, Andrej and Khosla, Aditya and Bernstein, Michael and others},
  journal={International journal of computer vision},
  volume={115},
  number={3},
  pages={211--252},
  year={2015},
  publisher={Springer}
}

@article{krizhevsky2009learning,
  title={Learning multiple layers of features from tiny images},
  author={Krizhevsky, Alex and Hinton, Geoffrey and others},
  year={2009},
  publisher={Toronto, ON, Canada}
}

@inproceedings{feki2025empirical,
  title={An empirical study on Hugging Face trends, topics and challenges on stack overflow},
  author={Feki, Hatem and Abdellatif, Manel and Sayagh, Mohammed},
  booktitle={2025 IEEE 49th Annual Computers, Software, and Applications Conference (COMPSAC)},
  pages={1297--1307},
  year={2025},
  organization={IEEE}
}

@phdthesis{mechalkhexploring,
  title={Exploring the Use of Large Language Models for Lossless Text Compression},
  author={Mechalkh, Charaf Eddine and Fennouh, Marya Douniazad},
  school={UNIVERSITY OF KASDI MERBAH OUARGLA}
}

@article{roumeliotis2023chatgpt,
  title={Chatgpt and open-ai models: A preliminary review},
  author={Roumeliotis, Konstantinos I and Tselikas, Nikolaos D},
  journal={Future Internet},
  volume={15},
  number={6},
  pages={192},
  year={2023},
  publisher={MDPI}
}

@inproceedings{morillo2026scalable,
  title={Scalable Microservices for LLM-vs-LLM Interaction in Board Games.},
  author={Morillo, Paulina and Bastidas, Kevin and Guevara, Bryan and Terreros, Alex and Proa{\~n}o, Julio},
  booktitle={MODELSWARD},
  pages={300--306},
  year={2026}
}

@inproceedings{lorentz2022profiling,
  title={Profiling the real world potential of neural network compression},
  author={Lorentz, Joe and Moawad, Assaad and Hartmann, Thomas and Aouada, Djamila},
  booktitle={2022 IEEE International Conference on Omni-layer Intelligent Systems (COINS)},
  pages={1--6},
  year={2022},
  organization={IEEE}
}

@inproceedings{hundt2026xprof,
  title={XProf: An Open, Scalable, and Extensible Profiling System for the Modern ML Stack},
  author={Hundt, Robert and Kumar, Naveen and Paredes, Jose Baiocchi and Goodson, Scott and Verghese, Clive and Rengasamy, Prasanna and Le, Kelvin and Zhang, Jiya and Alaras, Charles and Zhang, Yin and others},
  booktitle={Ninth Conference on Machine Learning and Systems},
  year={2026}
}

@inproceedings{Li_2020,
   title={XSP: Across-Stack Profiling and Analysis of Machine Learning Models on GPUs},
   url={http://dx.doi.org/10.1109/IPDPS47924.2020.00042},
   DOI={10.1109/ipdps47924.2020.00042},
   booktitle={2020 IEEE International Parallel and Distributed Processing Symposium (IPDPS)},
   publisher={IEEE},
   author={Li, Cheng and Dakkak, Abdul and Xiong, Jinjun and Wei, Wei and Xu, Lingjie and Hwu, Wen-mei},
   year={2020},
   month=May, pages={326–327} }

@techreport{hunger2005floating,
  title={Floating point operations in matrix-vector calculus.},
  author={Hunger, Raphael},
  year={2005},
  institution={Associate Institute for Signal Processing}
}

@inproceedings{qu2025automatic,
  title={Automatic joint structured pruning and quantization for efficient neural network training and compression},
  author={Qu, Xiaoyi and Aponte, David and Banbury, Colby and Robinson, Daniel P and Ding, Tianyu and Koishida, Kazuhito and Zharkov, Ilya and Chen, Tianyi},
  booktitle={Proceedings of the Computer Vision and Pattern Recognition Conference},
  pages={15234--15244},
  year={2025}
}

@article{van2020bayesian,
  title={Bayesian bits: Unifying quantization and pruning},
  author={Van Baalen, Mart and Louizos, Christos and Nagel, Markus and Amjad, Rana Ali and Wang, Ying and Blankevoort, Tijmen and Welling, Max},
  journal={Advances in neural information processing systems},
  volume={33},
  pages={5741--5752},
  year={2020}
}

@article{jafari2025profilingagent,
  title={ProfilingAgent: Profiling-Guided Agentic Reasoning for Adaptive Model Optimization},
  author={Jafari, Sadegh and Sarkar, Aishwarya and Bilwal, Mohiuddin and Jannesari, Ali},
  journal={arXiv preprint arXiv:2509.05584},
  year={2025}
}

@article{kodathala2026llms,
  title={LLMs can Compress LLMs: Adaptive Pruning by Agents},
  author={Kodathala, Sai Varun and Vunnam, Rakesh},
  journal={arXiv preprint arXiv:2601.09694},
  year={2026}
}

@inproceedings{devlin2019bert,
  title={Bert: Pre-training of deep bidirectional transformers for language understanding},
  author={Devlin, Jacob and Chang, Ming-Wei and Lee, Kenton and Toutanova, Kristina},
  booktitle={Proceedings of the 2019 conference of the North American chapter of the association for computational linguistics: human language technologies, volume 1 (long and short papers)},
  pages={4171--4186},
  year={2019}
}

@article{rangarajan2025identification,
  title={Identification of plant-parasitic nematode genera in turfgrass using deep learning algorithms},
  author={Rangarajan, Vikram and Shahoveisi, Fereshteh and Waldo, Benjamin D and Jafari, Sadegh},
  journal={Scientific Reports},
  year={2025},
  publisher={Nature Publishing Group UK London}
}

@article{kamilaris2018deep,
  title={Deep learning in agriculture: A survey},
  author={Kamilaris, Andreas and Prenafeta-Bold{\'u}, Francesc X},
  journal={Computers and electronics in agriculture},
  volume={147},
  pages={70--90},
  year={2018},
  publisher={Elsevier}
}

@article{litjens2017survey,
  title={A survey on deep learning in medical image analysis},
  author={Litjens, Geert and Kooi, Thijs and Bejnordi, Babak Ehteshami and Setio, Arnaud Arindra Adiyoso and Ciompi, Francesco and Ghafoorian, Mohsen and Van Der Laak, Jeroen Awm and Van Ginneken, Bram and S{\'a}nchez, Clara I},
  journal={Medical image analysis},
  volume={42},
  pages={60--88},
  year={2017},
  publisher={Elsevier}
}

@article{shuvo2022efficient,
  title     = {Efficient Acceleration of Deep Learning Inference on
               Resource-Constrained Edge Devices: A Review},
  author    = {Shuvo, Md Maruf Hossain and Islam, Syed Kamrul and
               Cheng, Jianlin and Morshed, Bashir I.},
  journal   = {Proceedings of the IEEE},
  volume    = {111},
  number    = {1},
  pages     = {42--91},
  year      = {2022},
  publisher = {IEEE}
}

@inproceedings{hinton2015distilling,
  title     = {Distilling the Knowledge in a Neural Network},
  author    = {Hinton, Geoffrey and Vinyals, Oriol and Dean, Jeff},
  booktitle = {NIPS Deep Learning and Representation Learning Workshop},
  year      = {2015}
}

@inproceedings{fang2023depgraph,
  title={Depgraph: Towards any structural pruning},
  author={Fang, Gongfan and Ma, Xinyin and Song, Mingli and Mi, Michael Bi and Wang, Xinchao},
  booktitle={Proceedings of the IEEE/CVF conference on computer vision and pattern recognition},
  pages={16091--16101},
  year={2023}
}

@inproceedings{dong2019hawq,
  title     = {{HAWQ}: Hessian {AW}are Quantization of Neural Networks with
               Mixed-Precision},
  author    = {Dong, Zhen and Yao, Zhewei and Gholami, Amir and
               Mahoney, Michael W. and Keutzer, Kurt},
  booktitle = {Proceedings of the IEEE/CVF International Conference on Computer
               Vision (ICCV)},
  pages     = {293--302},
  year      = {2019}
}

@article{hu2022dpro,
  title={dpro: A generic performance diagnosis and optimization toolkit for expediting distributed dnn training},
  author={Hu, Hanpeng and Jiang, Chenyu and Zhong, Yuchen and Peng, Yanghua and Wu, Chuan and Zhu, Yibo and Lin, Haibin and Guo, Chuanxiong},
  journal={Proceedings of Machine Learning and Systems},
  volume={4},
  pages={623--637},
  year={2022}
}

@misc{pytorch_profiler,
  author       = {{PyTorch Team}},
  title        = {PyTorch Profiler},
  year         = {2024},
  howpublished = {\url{https://pytorch.org/docs/stable/profiler.html}},
  note         = {Accessed: 2025-05-09}
}

@misc{tensorflow_profiler,
  author       = {{TensorFlow Team}},
  title        = {TensorFlow Profiler Guide},
  year         = {2024},
  howpublished = {\url{https://www.tensorflow.org/tensorboard/tensorboard_profiling_keras}},
  note         = {Accessed: 2025-05-03}
}

@misc{nvidia_profiler,
  author       = {{NVIDIA Corporation}},
  title        = {{CUDA Profiler User's Guide}},
  year         = {2024},
  howpublished = {\url{https://docs.nvidia.com/cuda/profiler-users-guide/index.html}},
  note         = {Accessed: 2025-05-03}
}

@misc{nvidia_cupti,
  author       = {{NVIDIA Corporation}},
  title        = {CUDA Profiling Tools Interface (CUPTI) Documentation},
  year         = {2024},
  howpublished = {\url{https://docs.nvidia.com/cupti/index.html}},
  note         = {Accessed: 2025-05-03}
}

@article{li2016ternary,
  title={Ternary weight networks},
  author={Li, Fengfu and Liu, Bin and Wang, Xiaoxing and Zhang, Bo and Yan, Junchi},
  journal={arXiv preprint arXiv:1605.04711},
  year={2016}
}

@article{shayer2017learning,
  title={Learning discrete weights using the local reparameterization trick},
  author={Shayer, Oran and Levi, Dan and Fetaya, Ethan},
  journal={arXiv preprint arXiv:1710.07739},
  year={2017}
}

@article{louizos2018relaxed,
  title={Relaxed quantization for discretized neural networks},
  author={Louizos, Christos and Reisser, Matthias and Blankevoort, Tijmen and Gavves, Efstratios and Welling, Max},
  journal={arXiv preprint arXiv:1810.01875},
  year={2018}
}

@article{wu2018training,
  title={Training and inference with integers in deep neural networks},
  author={Wu, Shuang and Li, Guoqi and Chen, Feng and Shi, Luping},
  journal={arXiv preprint arXiv:1802.04680},
  year={2018}
}

@article{uhlich2019mixed,
  title={Mixed precision dnns: All you need is a good parametrization},
  author={Uhlich, Stefan and Mauch, Lukas and Cardinaux, Fabien and Yoshiyama, Kazuki and Garcia, Javier Alonso and Tiedemann, Stephen and Kemp, Thomas and Nakamura, Akira},
  journal={arXiv preprint arXiv:1905.11452},
  year={2019}
}

@inproceedings{wang2020differentiable,
  title={Differentiable joint pruning and quantization for hardware efficiency},
  author={Wang, Ying and Lu, Yadong and Blankevoort, Tijmen},
  booktitle={European Conference on Computer Vision},
  pages={259--277},
  year={2020},
  organization={Springer}
}

@software{brevitas,
  author       = {Franco, Giuseppe and Monteagudo-Lago, Pablo and Colbert, Ian and Pappalardo, Alessandro and Fraser, Nicholas J},
  title        = {Xilinx/brevitas},
  year         = {2026},
  publisher    = {Zenodo},
  doi          = {10.5281/zenodo.3333552},
  url          = {https://doi.org/10.5281/zenodo.3333552}
}

@article{wei2022chain,
  title={Chain-of-thought prompting elicits reasoning in large language models},
  author={Wei, Jason and Wang, Xuezhi and Schuurmans, Dale and Bosma, Maarten and Xia, Fei and Chi, Ed and Le, Quoc V and Zhou, Denny and others},
  journal={Advances in neural information processing systems},
  volume={35},
  pages={24824--24837},
  year={2022}
}

@article{madaan2022text,
  title={Text and patterns: For effective chain of thought, it takes two to tango},
  author={Madaan, Aman and Yazdanbakhsh, Amir},
  journal={arXiv preprint arXiv:2209.07686},
  year={2022}
}

@inproceedings{yao2023react,
  title={React: Synergizing reasoning and acting in language models},
  author={Yao, Shunyu and Zhao, Jeffrey and Yu, Dian and Du, Nan and Shafran, Izhak and Narasimhan, Karthik and Cao, Yuan},
  booktitle={International Conference on Learning Representations (ICLR)},
  year={2023}
}

@article{zelikman2022star,
  title={Star: Bootstrapping reasoning with reasoning},
  author={Zelikman, Eric and Wu, Yuhuai and Mu, Jesse and Goodman, Noah},
  journal={Advances in Neural Information Processing Systems},
  volume={35},
  pages={15476--15488},
  year={2022}
}

@article{ahn2022can,
  title={Do as i can, not as i say: Grounding language in robotic affordances},
  author={Ahn, Michael and Brohan, Anthony and Brown, Noah and Chebotar, Yevgen and Cortes, Omar and David, Byron and Finn, Chelsea and Fu, Chuyuan and Gopalakrishnan, Keerthana and Hausman, Karol and others},
  journal={arXiv preprint arXiv:2204.01691},
  year={2022}
}

@article{lin2023swiftsage,
  title={Swiftsage: A generative agent with fast and slow thinking for complex interactive tasks},
  author={Lin, Bill Yuchen and Fu, Yicheng and Yang, Karina and Brahman, Faeze and Huang, Shiyu and Bhagavatula, Chandra and Ammanabrolu, Prithviraj and Choi, Yejin and Ren, Xiang},
  journal={Advances in Neural Information Processing Systems},
  volume={36},
  pages={23813--23825},
  year={2023}
}

@inproceedings{park2023generative,
  title={Generative agents: Interactive simulacra of human behavior},
  author={Park, Joon Sung and O'Brien, Joseph and Cai, Carrie Jun and Morris, Meredith Ringel and Liang, Percy and Bernstein, Michael S},
  booktitle={Proceedings of the 36th annual acm symposium on user interface software and technology},
  pages={1--22},
  year={2023}
}

@article{schick2023toolformer,
  title={Toolformer: Language models can teach themselves to use tools},
  author={Schick, Timo and Dwivedi-Yu, Jane and Dess{\`\i}, Roberto and Raileanu, Roberta and Lomeli, Maria and Hambro, Eric and Zettlemoyer, Luke and Cancedda, Nicola and Scialom, Thomas},
  journal={Advances in Neural Information Processing Systems},
  volume={36},
  pages={68539--68551},
  year={2023}
}

@online{ptflops,
  author = {Vladislav Sovrasov},
  title = {ptflops: a flops counting tool for neural networks in pytorch framework},
  year = {2018-2024},
  url = {https://github.com/sovrasov/flops-counter.pytorch},
}

@article{kullback1951information,
  title={On information and sufficiency},
  author={Kullback, Solomon and Leibler, Richard A},
  journal={The annals of mathematical statistics},
  volume={22},
  number={1},
  pages={79--86},
  year={1951},
  publisher={JSTOR}
}

\end{document}